\documentclass[11pt]{article}

\usepackage[final]{acl}

\usepackage{times}
\usepackage{latexsym}

\usepackage[T1]{fontenc}

\usepackage[utf8]{inputenc}

\usepackage{microtype}

\usepackage{inconsolata}

\usepackage{amsmath}
\usepackage{amsfonts}
\usepackage{amssymb}
\usepackage{graphicx}
\usepackage{hyperref}
\usepackage{algorithm}
\usepackage{algorithmic}
\usepackage{booktabs}
\usepackage{subcaption}
\usepackage{xcolor}
\usepackage{colortbl}
\usepackage{tcolorbox}
\usepackage{float}

\title{Towards Efficient Reasoning: Learning Causal Shortcuts for Diffusion Language Models}

\author{
\textbf{Dian Jin}$^{1}$\thanks{Equal contribution.},
\textbf{Kairong Han}$^{1}$\footnotemark[1],
\textbf{Baohong Li}$^{1}$,
\textbf{Xinpeng Dong}$^{1}$,
\\
\textbf{Zijing Hu}$^{1}$,
\textbf{Nuanqiao Shan}$^{1}$,
\textbf{Fei Wu}$^{1,2}$,
\textbf{Kun Kuang}$^{1}$\thanks{Corresponding author.}
\\
$^{1}$Zhejiang University,
$^{2}$Shanghai AI Laboratory
\\
\href{mailto:dianjin@zju.edu.cn}
{\texttt{dianjin@zju.edu.cn}}
\qquad
\href{mailto:kunkuang@zju.edu.cn}
{\texttt{kunkuang@zju.edu.cn}}
}

\begin{document}
\maketitle


\begin{abstract}
Diffusion Language Models (DLMs) have attracted significant attention for their strong reasoning ability. However, under a bidirectional attention mechanism, DLMs operate over an exponentially large exploration space compared to autoregressive models (ARMs), making it challenging to focus on reasoning-guiding tokens under random masking. We define causal shortcuts as token chains that cover the full sequence and provide explicit guidance towards correct reasoning trajectories. We analyze the effects of causal shortcuts on the reasoning accuracy and convergence speed of DLMs, and find that they largely improve answer convergence efficiency and generation accuracy. Motivated by this, we propose a Causal Shortcut Learning  (CSL) Framework for DLMs. Specifically, we introduce a step-by-step token extraction procedure to extract causal shortcuts from data, and apply parallel prioritized masking on these tokens during training to enable efficient and accurate convergence to correct answers via causal shortcuts. Extensive experiments across multiple reasoning benchmarks and two base models demonstrate that CSL consistently outperforms existing SFT-variant baselines, achieving an average improvement of $1.92\%$ over SFT-only models, and up to $4.20\%$ on MATH-500. The code is available at the \href{https://github.com/ZJUDianJin/Causal-Shortcuts-Learning}{https://github.com/ZJUDianJin/Causal-Shortcuts-Learning}.

\end{abstract}

\begin{figure}[t]
    \centering
    \includegraphics[width=1.05\linewidth, trim=2mm 0mm 0mm 0mm, clip]{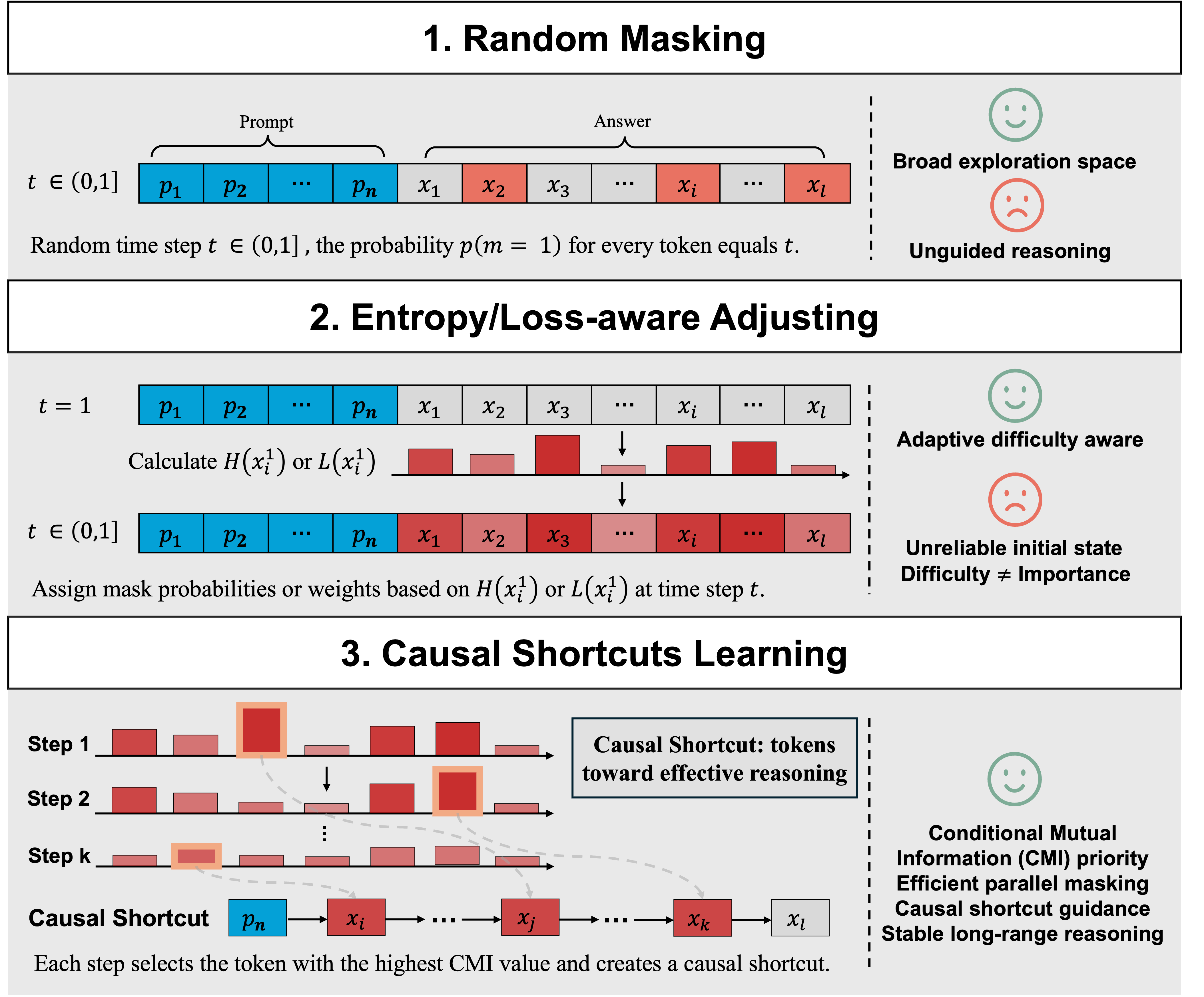}
    \caption{Comparison of three SFT paradigms: Random Masking, entropy/loss-aware adjusting, CSL.}
    \label{fig:intro}
\end{figure}

\section{Introduction}

Recently, diffusion-based large language models have gained increasing attention. In particular, Diffusion Language Models (DLMs) demonstrate strong performance and reasoning ability \cite{llada, ou2025your, yang2026mmada}. With bidirectional-attention mechanisms and flexible generation paradigms, DLMs replace strictly autoregressive sequential decoding with efficient parallel generation and provide a new perspective and promising direction for LLM development \cite{ye2025dream, ye2025beyond, gong2025scaling, han2025c}.

During supervised fine-tuning, autoregressive models (ARMs) learn from strictly deterministic left-to-right generation trajectories \cite{vaswani2017attention, radford2018improving, radford2019language, brown2020language}. In contrast, DLMs replace this paradigm with random masking and bidirectional attention, enabling access to a richer set of generation trajectories \cite{kim2025train, gisserot2025should}. This exponentially expands the trajectory space. While this theoretically endows DLMs with greater reasoning potential \cite{svete2025reasoning}, the exponentially expanded space makes it difficult for the model to focus on reasoning-guiding tokens, leading it to prioritize simple, high-frequency ones under random masking.

Existing works have explored the importance heterogeneity among tokens in DLM training by reweighting tokens based on entropy and loss under full mask sequence \cite{mgdm}, or by adjusting masking probabilities \cite{db, dsft, gift}. However, these approaches largely equate "difficulty" or "uncertainty" with importance. This perspective is insufficient to capture true reasoning guidance. For example, tokens that are simple but distant from the prompt may exhibit high uncertainty, causing the model to over-focus on them. Moreover, tokens with different parts of speech exhibit different entropy and loss behaviors.


How can we effectively define token importance? We propose a conditional mutual information (CMI) score to quantify the causal information that a token contributes to the rest of the sequence. Specifically, while holding the current masked context fixed, we reveal a token and measure the resulting reduction in the entropy of the remaining masked tokens. The causal direction is established through this reveal-and-measure process: the revealed token serves as the information source, and CMI quantifies its explanatory contribution to the remaining masked tokens by measuring the reduction in their entropy. We extract high-CMI tokens and assemble them into token chains, which we term causal shortcuts. These shortcuts span the entire sequence and provide strong guidance toward correct reasoning trajectories. When used as prompts, they enable DLMs to produce complete answers in fewer decoding steps and yield substantial accuracy gains, as shown in Figure~\ref{fig:shortcut_all}. These results suggest that causal shortcuts provide shorter reasoning paths that guide the model toward correct trajectories.

Motivated by the strong guiding effect of causal shortcuts, we propose a \textbf{C}ausal \textbf{S}hortcut \textbf{L}earning (CSL) Framework for DLMs. We first introduce a step-by-step token extraction procedure to extract causal shortcuts from data. To encourage DLMs to focus more on them, we adopt a simple and effective strategy: apply parallel prioritized masking on causal shortcuts during training to enable efficient and accurate convergence to correct answers via causal shortcuts. This strategy is motivated by two considerations: (1) causal shortcut tokens require richer exploration space for effective learning; therefore, parallel masking enables efficient gradient updates, making DLMs largely focus on them; (2) causal shortcut tokens exhibit weak local dependencies, ensuring that parallel masking does not disrupt dependencies.

We evaluate CSL on mathematical reasoning and code generation tasks. Built upon LLaDA-8B-Instruct and LLaDA-1.5, our method outperforms the original model, SFT-only model and five SFT variant baselines. It achieves an average improvement of $1.92\%$ across seven math benchmarks and $2.30\%$ across two code benchmarks, with a prominent gain of $4.20\%$ on MATH-500. Ablations show that CSL leads to faster entropy decay and reduced cumulative entropy, facilitating effective reasoning and mitigating error accumulation. Our contributions can be summarized as follows:

\begin{itemize}
\item We propose the Conditional Mutual Information (CMI) score to quantify each token's impact on reasoning, and further propose causal shortcuts based on CMI, which effectively guide the model toward correct reasoning trajectories.

\item We propose a Causal Shortcut Learning framework (CSL) for DLMs, which applies parallel masking to encourage the model to focus more on causal shortcuts, enabling more efficient and accurate reasoning.

\item Extensive experiments across mathematical and coding domains demonstrate that CSL achieves state-of-the-art performance, consistently outperforming existing baselines.
\end{itemize}

\section{Preliminary and Related Works}
\subsection{Diffusion Language Models}
DLM is a masked diffusion-based language model, whose forward process progressively corrupts the original sequence by token masking \cite{austin2021structured, sahoo2024simple, llada, lou2023discrete}. At a given time step $t \in (0,1]$, the original sequence is transformed into a noised version $x_t$, where a subset of tokens is masked. As the time step increases, the probability of each token remaining unmasked $\alpha_t$ decreases monotonically, leading to a progressively more corrupted sequence, until all tokens are masked at $t = 1$.

DLMs employ a bidirectional attention mechanism to model the conditional distribution over masked tokens. At each training step, a time step $t \in [0,1)$ is sampled and a corrupted sequence is constructed via a forward masking process. The model then predicts all masked tokens in parallel. DLMs are optimized using the Negative Evidence Lower Bound (NELBO), which serves as an upper bound on the negative log-likelihood (NLL) \cite{llada}.

In this work, we follow the LLaDA series and adopt its NELBO formulation with a linear noise schedule $\alpha_t = 1 - t$. The corresponding NELBO objective is defined as:

\begin{equation*}
- \mathbb{E}_{t, p_0, x_0, x_t}
  \left[ \frac{1}{t}
  \sum_{i=1}^{\mid x_t \mid}
  \mathbb{I}[x_t^i = \text{M}]
  \cdot \log p_{\theta}(x_0^i \mid p_0, x_t)
  \right],
\end{equation*}

\noindent where $x_t$ denotes the corrupted sequence with mask noise, $|x_t|$ denotes the length of the sequence, and $p_0$ denotes the unmasked prompt.

\subsection{Training and Masking Strategies}
Recent works have attempted to improve DLMs' reasoning ability via different perspectives. MGDM \cite{mgdm} reweight tokens based on their loss in $x_{t=1}$ by emphasizing those with larger loss values to improve optimization. DiffusionBert \cite{db} and GIFT \cite{gift} assign different masking probabilities based on token-level entropy. Blockwise \cite{blockwise} introduces blockwise masking in a semi-autoregressive manner. DSFT \cite{dsft} further adjusts masking probabilities and weights for numerical tokens, and incorporates span-level and curriculum masking strategies. The methods above are detailed in Appendix~\ref{appendix:baselines}.

Unlike most approaches equating "difficulty" or "uncertainty" with importance, CSL takes a causal perspective to encourage the model to focus on strongly reasoning-guiding tokens, enabling more efficient reasoning.
\section{Method}

\begin{figure*}[t]
\centering
\begin{subfigure}{0.58\textwidth}
    \centering
    \includegraphics[width=1.0\textwidth]{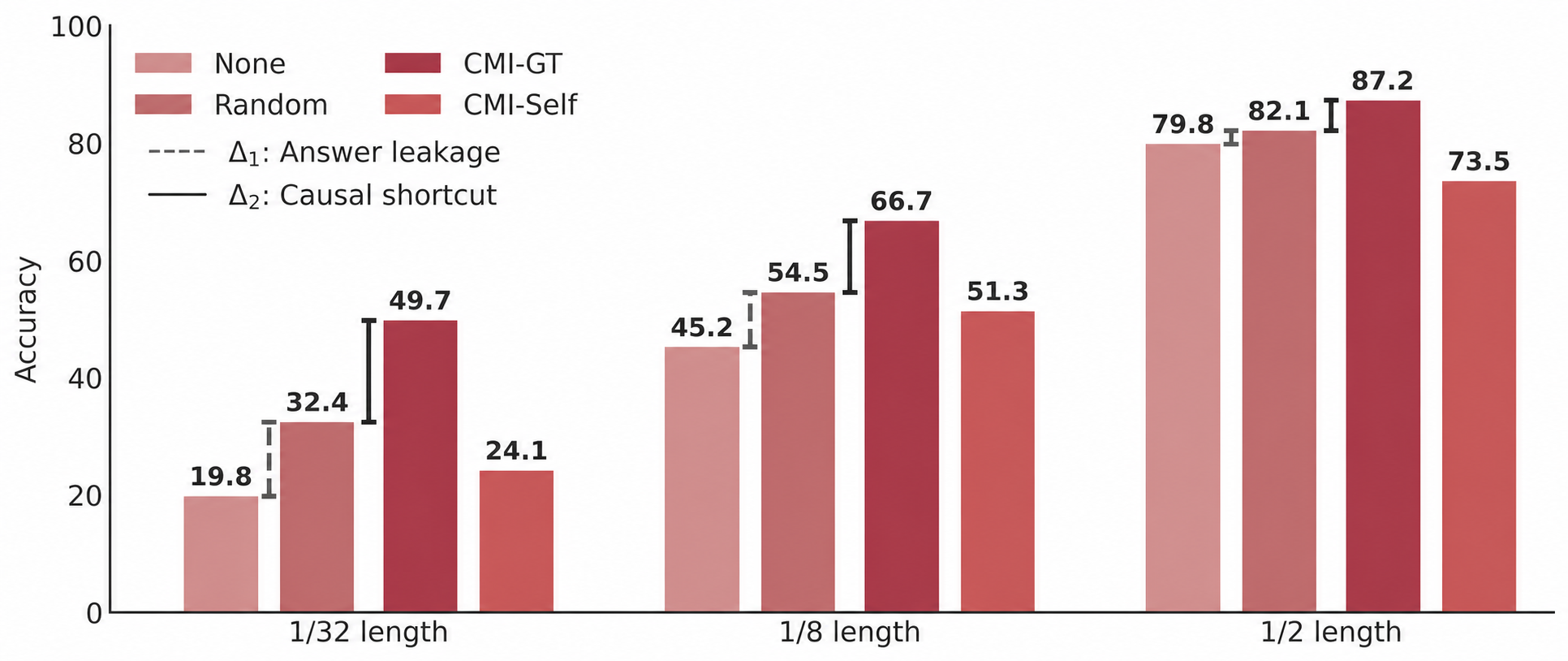}
    \caption{}
    \label{fig:shortcut}
\end{subfigure}
\hfill
\begin{subfigure}{0.38\textwidth}
    \centering
    \includegraphics[width=1.0\textwidth]{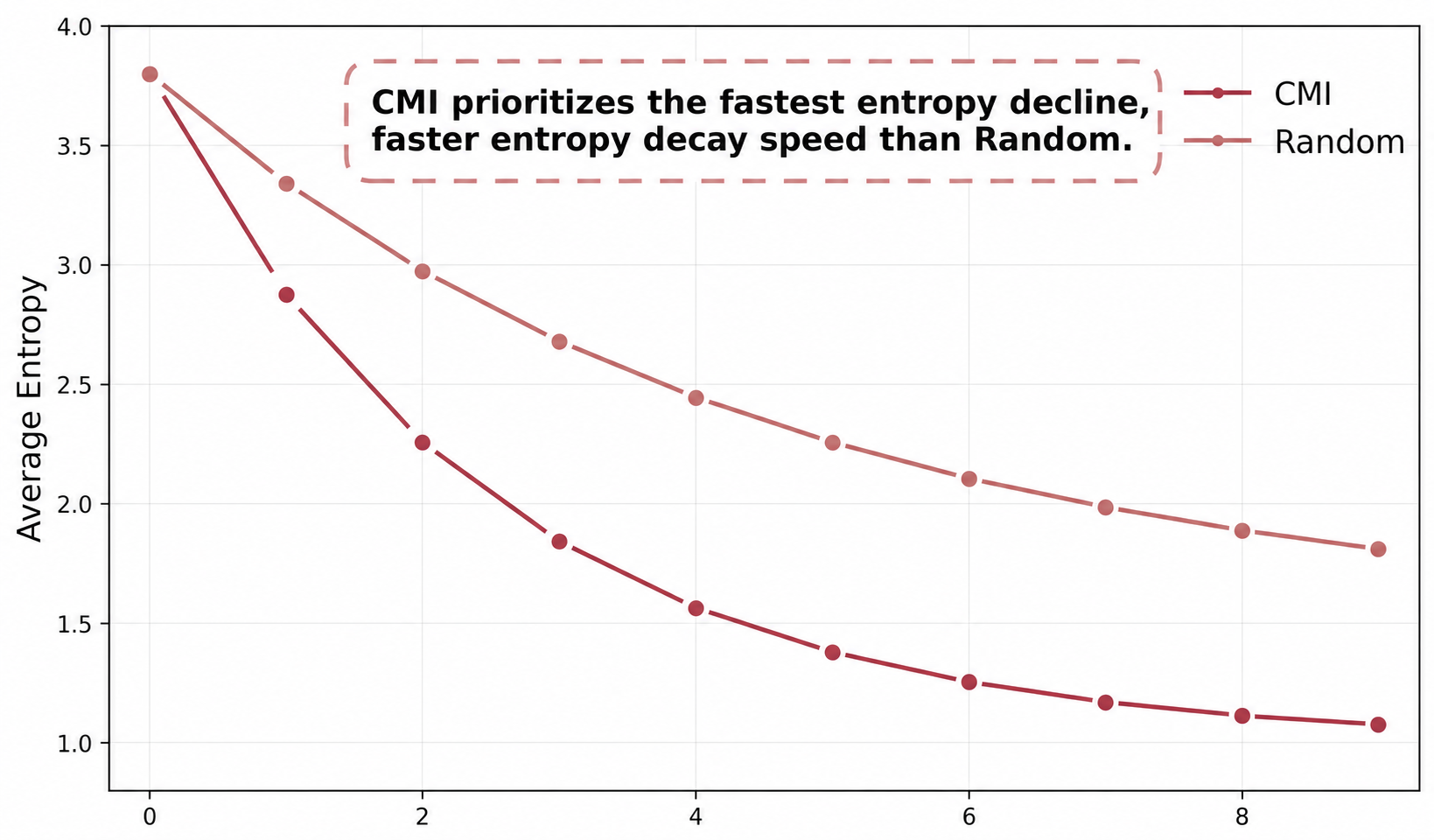}
    \caption{}
    \label{fig:entropy_polyline}
\end{subfigure}
\caption{
(a) We compare four token selection strategies for reasoning accuracy: 1.None; 2.Random; 3.CMI-GT, tokens are selected via CMI using ground-truth answers; 4.CMI-Self, using model-sampled tokens. $\Delta_1$ reflects gains from answer leakage, while $\Delta_2$ reflects improvements from causal trajectory guidance. (b) Entropy decay comparison between CMI-GT and Random. CMI-GT induces faster entropy reduction and more efficient reasoning convergence.
}
\label{fig:shortcut_all}
\end{figure*}

\subsection{Conditional Mutual Information Priority}
Autoregressive models (ARMs) employ causal masking, restricting generation to a left-to-right factorization. In contrast, DLMs shift this paradigm towards global sequence modeling, enabling each token to condition on arbitrary masked contexts. Consequently, DLMs operate over an exponentially large exploration space, making it difficult to efficiently identify effective reasoning trajectories and improve generation accuracy.

This raises a natural question: in such an exponentially large exploration space, are there key reasoning-guiding tokens that can serve as "shortcuts" toward correct answers? Entropy provides a natural measure of uncertainty. We interpret the entropy of the token distribution at each masked position as its information uncertainty. Formally, we define the entropy at position $i$ as:

\begin{equation*}
H(x_t^i)= - \sum_{v \in \mathcal{V}}p_{\theta}(x_t^i = v \mid x_t)\log p_\theta (x_t^i = v \mid x_t)
\end{equation*}

\noindent where $\mathcal{V}$ denotes the vocabulary. Formally, we use conditional mutual information (CMI) to quantify the information that a token contributes to the remaining unmasked sequence. The CMI score is defined as follows:

\begin{equation*}
\text{CMI}(x_t^i)
=
\frac{1}{|\mathcal{M}_t|-1}
\sum_{j \in \mathcal{M}_t \setminus \{i\}}
H(x_t^j)
-
H(x_t^j \mid x_0^i)
\end{equation*}

\noindent where $\mathcal{M}_t$ denotes the set of masked indices in the noisy sequence $x_t$. Each term can be interpreted as the conditional mutual information \cite{kraskov2004estimating} between the revealed token $x_i^0$ and another masked token $x_j^t$ under the current noisy state:

\begin{equation*}
I(x_0^i ; x_t^j \mid x_t)
=
H(x_t^j)
-
H(x_t^j \mid x_0^i)
\end{equation*}

\noindent This formulation measures the uncertainty reduction of $x_j^t$ after revealing $x_i^0$, which corresponds to the conditional mutual information between $x_i^0$ and $x_j^t$ under the noisy state $x_t$. Therefore, the proposed conditional mutual information (CMI) can be interpreted as the expected pairwise conditional mutual information between token $x_i^0$ and other masked tokens:

\begin{equation*}
\text{CMI}(x_t^i)
=
\mathbb{E}_{j \sim \mathcal{M}_t \setminus \{i\}}
\left[
I(x_0^i ; x_t^j \mid x_t)
\right]
\end{equation*}

\noindent Thus, CMI characterizes the dependency between $x_i^0$ and the current masked sequence $x_t$, reflecting the importance of $x_i^0$ as a key node in reasoning trajectories.


\noindent \textbf{Experiments on CMI Priority.} We select causal shortcut tokens based on conditional mutual information (CMI) and incorporate them into the input prompt together with the original question. We evaluate this design on mathematical generation tasks to investigate its effect on accuracy and entropy under different generation steps. And the results are shown in Figure \ref{fig:shortcut_all}.

\begin{figure*}[t]
    \centering
    \includegraphics[width=1.0\linewidth, trim=7mm 6mm 6mm 6mm, clip]{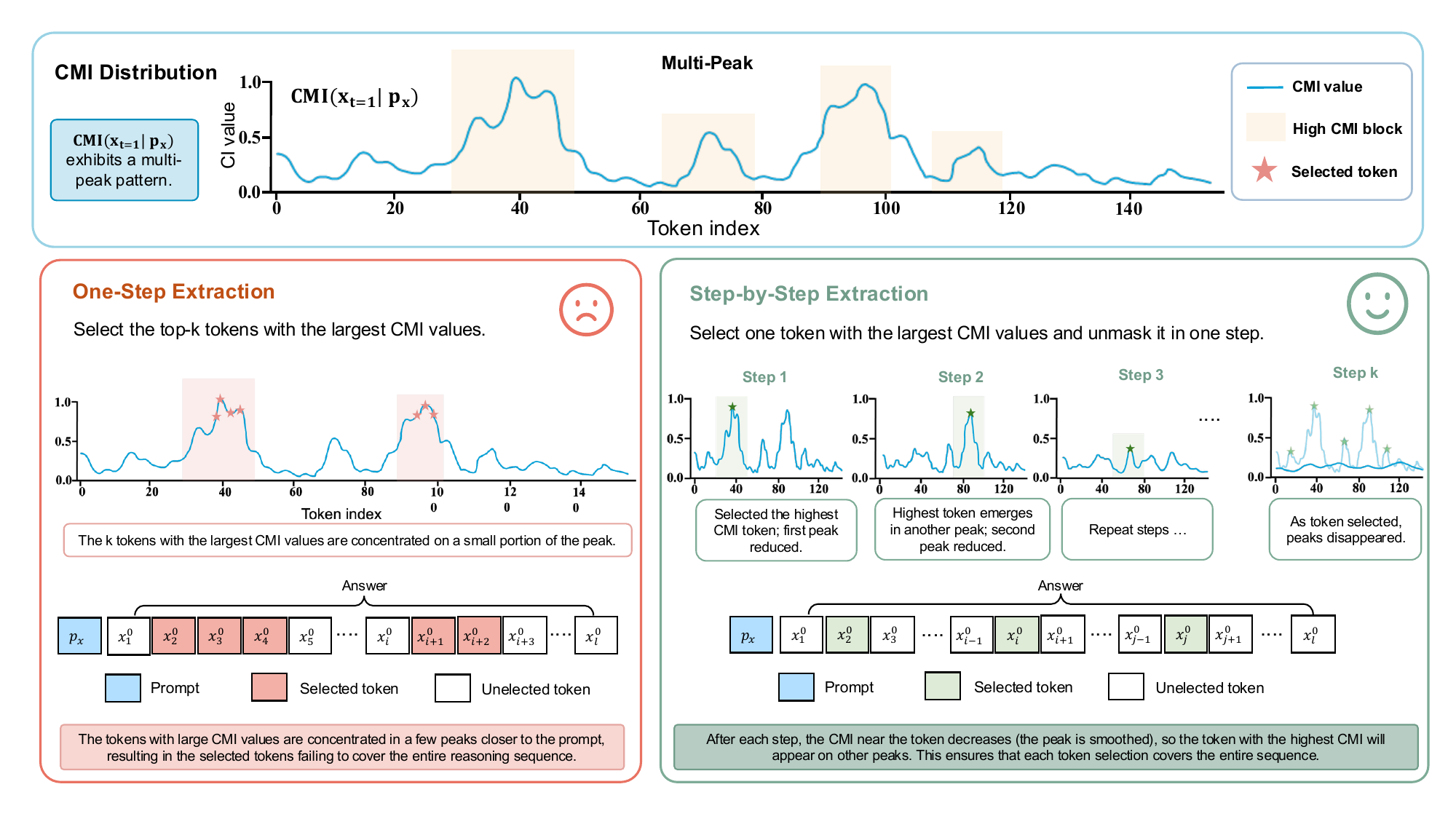}
    \caption{Comparison of two extraction strategies: (1) One-step extraction selects the top-$k$ CMI tokens from the fully masked sequence $x_{t=1}$, leading to high-CMI tokens cluster, failing to cover the full reasoning trajectory; (2) Step-by-step extraction iteratively selects and unmasks the highest-CMI token. Updated CMI distribution shifts high-value token to new region, connects multiple peaks into a coherent reasoning trajectory, forming a causal shortcut.}
    \label{fig:extraction}
\end{figure*}

\begin{tcolorbox}[colback=red!10,
    colframe=red!60!black,
    boxrule=0.8pt]
\textbf{Observation 1: Causal shortcuts substantially improve accuracy and efficiency.}
\end{tcolorbox}
We observe that this intervention significantly improves both accuracy and generation efficiency. As shown in Figure \ref{fig:shortcut}, we analyze performance under different decoding steps. Results show that CMI-GT substantially improves accuracy and reduces entropy, enabling more efficient convergence with fewer timesteps. Notably, under the 1/32-length timestep setting, CMI-GT achieves a remarkable $17.3\%$ improvement. Moreover, CMI-GT further raises the accuracy to $87.2\%$ under larger timestep settings. This improvement suggests that causal shortcuts provide strong guidance toward correct reasoning trajectories, forming an effective shortcut to the final answer.

\begin{tcolorbox}[colback=red!10,
    colframe=red!60!black,
    boxrule=0.8pt]
\textbf{Observation 2: Self-generated tokens accumulate errors and reduce accuracy.}
\end{tcolorbox}
However, it is extremely challenging for DLMs to directly generate such causal trajectories based on masked sequence $x_1$. Relative to the ground-truth, the tokens generated by DLMs are inherently unreliable, leading to incorrect reasoning trajectories. As shown in \ref{fig:shortcut}, CMI-Self leads to a substantial degradation in accuracy. We draw two conclusions as follows: (1) biased causal trajectories severely impair reasoning correctness; (2) DLMs suffer from error accumulation.

Motivated by two observations, we propose Causal Shortcut Learning (CSL) Framework, which extracts tokens based on CMI and encourages DLMs to focus on learning causal shortcuts during training.

\subsection{Causal Shortcuts Learning}

\subsubsection{One-Step Extraction}
One-step extraction directly obtains top-$k$ tokens with highest CMI scores via a single forward pass on the fully masked sequence $x_{t=1}$. Given a causal shortcut set size $K$, the one-step extraction can be formulated as:
$$
\mathcal{S}_{\text{one-step}} 
= \bigl\{ x_0^i \,\big|\, 
i\in\mathop{\arg\max}_{\substack{\mathcal{S}\subseteq\{1,\dots,L\}, |\mathcal{S}|=K}}
\sum_{j\in\mathcal{S}}\text{CMI}(x_0^j) \bigr\}
$$
As shown in Figure \ref{fig:extraction}, one-step extraction strategy leads to a clustered distribution of high-CMI tokens near the prompt, resulting in the selected tokens failing to cover the entire reasoning sequence. This phenomenon arises from two factors: (1) high-CMI tokens tend to form local clusters due to the inherent clustering structure of information; (2) low-CMI tokens emerge at distant positions, as prompt-induced autoregressive conditioning limits model perception of distant masked tokens. Causal Shortcuts consider both CMI value and distribution of extracted tokens.

\subsubsection{Step-by-step Extraction} 
To address this issue, we propose a step-by-step CMI-based token extraction methods. Instead of selecting all shortcut tokens at one step, we iteratively select the CMI-highest token and unmasked it to update CMI distribution. Let $\mathcal{S}_{0} = \emptyset$. Let $x_{k} = \{x^i_{k}\}_{i=1}^{L}$ denote the sequence at iteration $k$, where $x_{0}$ is the fully masked initial sequence. At iteration $k \in \{1,\dots,K\}$, we greedily add the selected token to the shortcut set:
$$
\mathcal{S}_{k}
=
\mathcal{S}_{k-1}
\cup
\left\{
\arg\max_{i \notin \mathcal{S}_{k-1}}
\text{CMI}(x^i_{k-1})
\right\}.
$$

\noindent After each iteration, we unmask the selected token and update CMI on sequence $x_{k}$. After $K$ iterations, the final causal shortcut set is
$$
\mathcal{S}_{\text{k-step}} = \mathcal{S}_{K}.
$$

This step-by-step extraction strategy alleviates the token clustering problem in one-step extraction. Once a high-CMI token is unmasked, local uncertainty decreases, leading to a reduction in CMI. As illustrated in Figure \ref{fig:extraction}, the initial CMI distribution exhibits multiple-peaks. The reduction of local uncertainty suppresses the CMI peaks for each unmasked token. Through step-by-step extraction, causal shortcuts form a reasoning trajectory that covers the entire sequence, yielding an efficient guidance for correct reasoning. We provide a theoretical analysis of this process in the Appendix~\ref{analysis}. And the pseudocode is provided in Algorithm~\ref{alg:token_extraction}.

\begin{table}[t]
\centering
\caption{
Complexity comparison of the global-search and sliding-window
extraction strategies. The simplified results assume
$K=\Theta(L)$ and $D=O(1)$.
}
\label{tab:extraction_complexity}
\resizebox{\columnwidth}{!}{
\begin{tabular}{lccc}
\toprule
Method
& Candidates/step
& Total complexity
& FP batches \\
\midrule
Global
& $L$
& $O(KL)=O(L^2)$
& $O\!\left(K\left\lceil L/B\right\rceil\right)$ \\
Window
& $D$
& $O(KD)=O(L)$
& $O\!\left(K\left\lceil D/B\right\rceil\right)$ \\
\bottomrule
\end{tabular}
}
\end{table}

\noindent \textbf{Time Complexity Optimization.}
As shown in Table~\ref{tab:extraction_complexity}, the naive step-by-step method searches over all $L$ candidate positions at each of the $K$ extraction steps, resulting in a complexity of $O(KL)$. Since $K=\Theta(L)$, this procedure scales quadratically with the sequence length. We empirically observe that high-CMI tokens are sparsely distributed and therefore replace the global search with a sliding-window search over $D$ candidates at each step. This reduces the complexity to $O(KD)$, which becomes $O(L)$ when $D$ is fixed independently of $L$. With batch size $B$, the number of forward-pass batches is further reduced to $O\!\left(K\left\lceil D/B\right\rceil\right)$.

\noindent \textbf{Score Model for CMI Prediction}
Although the sliding-window strategy reduces the asymptotic complexity of step-by-step extraction, applying it to every example in a large-scale dataset still incurs substantial preprocessing costs. To address this issue, we train a scoring model on a small data subset to estimate CMI on all data, replacing time-consuming calculations, significantly improves the efficiency. In experiments, the rank metric \text{NDCG} of CMI estimation reaches $0.92$. A detailed implementation is provided in Appendix \ref{appendix:scaling}.

\subsubsection{Parallel Masking Training}

We adopt a simple yet efficient strategy to encourage DLMs to focus on Causal Shortcuts via parallel masking. We define $\mathcal{S}$ as causal shortcuts set. The training objective is formulated as follows:

\begin{equation*}
\begin{aligned}
\mathcal{L}_{\text{CSL}}
=
-\mathbb{E}_{t, p_0, x_0, x_t, \mathcal{S}}
\Bigg[
\underbrace{
\frac{1}{t}\sum_{i \in \mathcal{S}}
\log p_{\theta}(x^i_0 \mid p_0, x_t)
}_{\text{parallel masking}(\mathcal{S})} \\
+
\underbrace{
\frac{1}{t}\sum_{i \notin \mathcal{S}}
\mathbb{I}[x_t^i = \text{M}]
\log p_{\theta}(x^i_0 \mid p_0, x_t)
}_{\text{random masking}}
\Bigg]
\end{aligned}
\end{equation*}

This strategy is motivated by two properties of extracted causal shortcut set. (1) Causal shortcuts cover the entire generation trajectory. Parallel masking enables more efficient gradient updates, making the model focus more on them and improving reasoning performance. (2) Causal shortcuts cover the entire reasoning trajectory and exhibit weak local dependencies, avoid parallel pitfalls caused by token accumulation: masking strong-dependent tokens simultaneously breaks inherent relations, hindering effective learning and degrading reasoning performance.


\section{Experiments}
\subsection{Experimental Setup}

\subsubsection{Benchmark}

We evaluate the effectiveness of CSL on both mathematical reasoning and code generation tasks. For mathematical reasoning, we conduct experiments on seven widely used benchmarks: (1) \textbf{GSM8K} \cite{gsm8k}; (2) \textbf{MATH-500} \cite{math-500}; (3) \textbf{SAT} \cite{sat}; (4) \textbf{Sudoku} \cite{d1}; (5) \textbf{GPQA} \cite{gpqa}; (6) \textbf{MMLU-STEM} \cite{mmlu}; (7) \textbf{ARC-C} \cite{arc}. For code generation, we evaluate CSL on two standard benchmarks: (1) \textbf{HumanEval} \cite{humaneval}; (2) \textbf{MBPP} \cite{mbpp}.

\subsubsection{Base Models}
We conduct training and evaluation based on two representative DLMs from the LLaDA series, including \textbf{LLaDA-8B-Instruct} \cite{llada} and \textbf{LLaDA-1.5B} \cite{llada1.5}. We apply CSL on top of these models to evaluate its effectiveness across diverse reasoning tasks.

\subsubsection{Baselines}
We compare CSL with several SFT-variants of diffusion-based models, including: (1) \textbf{DiBT} (DiffusionBert) \cite{db}; (2) \textbf{MGDM} \cite{mgdm}; (3) \textbf{Blockwise} \cite{blockwise}; (4) \textbf{DSFT} \cite{dsft}; (5) \textbf{GIFT} \cite{gift}. These baselines improve diffusion-based training paradigms from different perspectives to enhance reasoning capabilities.

\subsubsection{Datasets and Training}
We use two high-quality datasets from the math and code domains, \textbf{Math-CoT} \cite{math-cot} and \textbf{OPC-SFT-Stage2} \cite{opc}. We preprocess both datasets to extract causal shortcut sets, which are used as training data. For SFT, we train all base models under the same hyperparameter settings across different methods. We apply LoRA \cite{lora} for fine-tuning with a fixed learning rate of $2 \times 10^{-4}$. On mathematical reasoning tasks, we train for $4$ epochs on LLaDA-8B-Instruct and $8$ epochs on LLaDA-1.5B, with a sequence length of $2048$. On code generation tasks, we train for $4$ epochs on LLaDA-8B-Instruct, with a sequence length of $1024$. Detailed training configurations and hyperparameters are provided in Appendix \ref{appendix:hyper}.

\begin{figure}[t]
\centering
\includegraphics[width=1.0\linewidth]{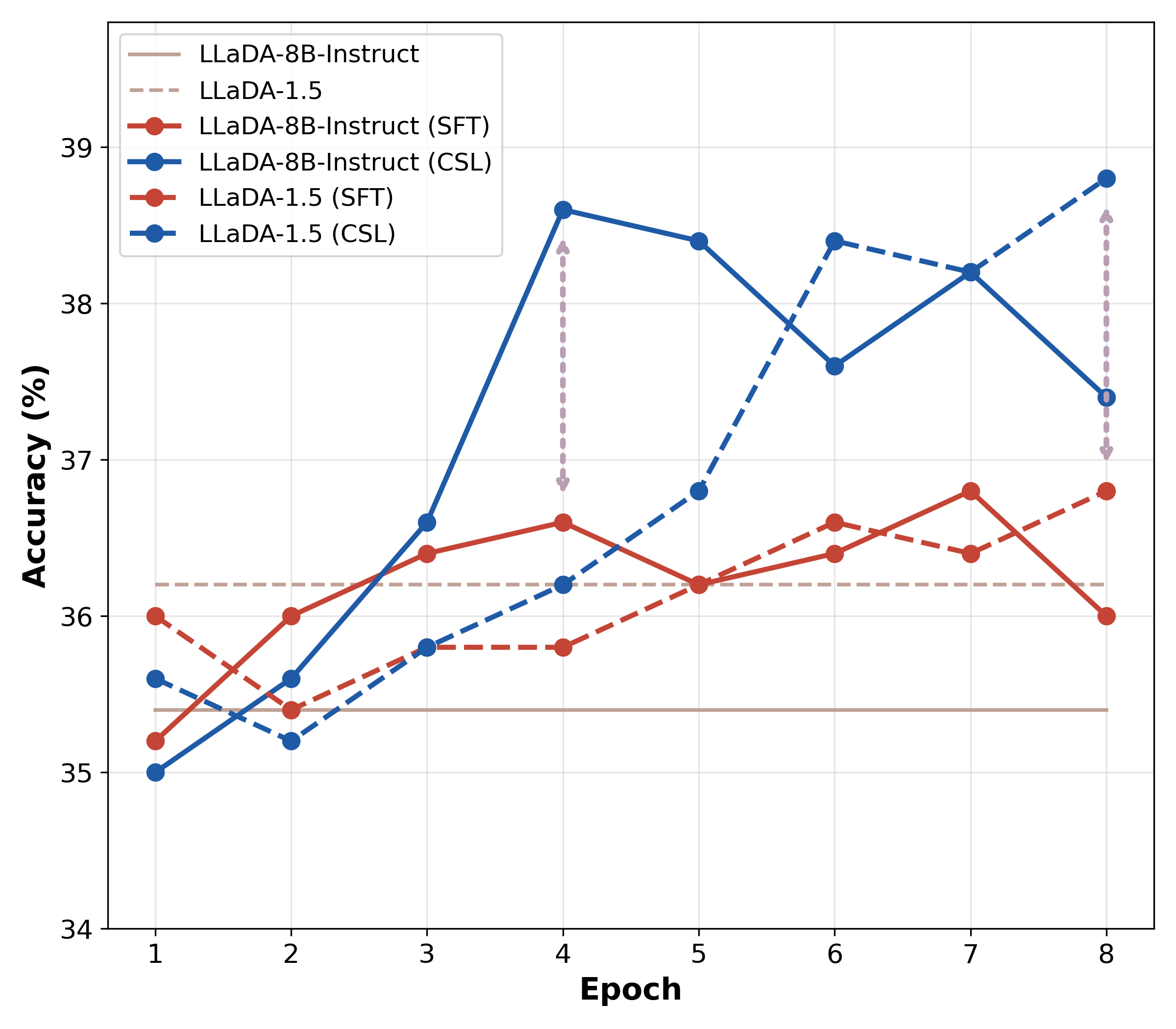}
\caption{Accuracy curve with training progress on the MATH-500 task for two base models.
}
\label{fig:acc_epochs}
\end{figure}

We investigate CSL performance across epoch checkpoints on MATH-500. As shown in Figure~\ref{fig:acc_epochs},    significant accuracy gains are observed at the 4th and 8th training epochs for the two base models, respectively. We further provide the analysis of training stability in the Appendix~\ref{appendix:training_loss}.

\renewcommand{\arraystretch}{1.20}
\begin{table*}[ht]
\centering
\small
\setlength{\tabcolsep}{2.8pt}
\caption{Performance comparison of CSL with SFT-based diffusion variants. Avg is computed as the mean over all nine tasks. $\Delta$ denotes absolute improvement over Base and SFT models, respectively. The best and second-best results are highlighted in \textbf{bold} and \underline{underlined}.}
\label{tab:results}
\begin{tabular}{l|ccccccccc|c}
\toprule
\bfseries
\textbf{Method} & \textbf{GSM8K-256} & \textbf{GSM8K-512} & \textbf{MATH-256} & \textbf{MATH-512} & \textbf{SAT} & \textbf{Sudoku} & \textbf{GPQA} & \textbf{MMLU} & \textbf{ARC-C} & \textbf{Avg} \\
\midrule

\rowcolor{gray!20}
\multicolumn{11}{c}{\textbf{LLaDA-8B-Instruct}} \\
\midrule

\textbf{Base} & 76.19\% & \underline{80.36\%} & 32.80\% & 35.40\% & 76.36\% & 11.18\% & 28.12\% & 61.21\% & 84.98\% & 54.07\% \\
\textbf{SFT}      & \underline{79.08\%} & 79.23\% & 32.20\% & 36.60\% & 75.91\% & 11.28\% & 28.57\% & 61.40\% & 85.49\% & \underline{54.41}\% \\
\midrule

DiBT & 78.24\% & 79.15\% & 35.00\% & \underline{37.00\%} & 76.36\% & 5.05\%  & 28.57\% & 61.15\% & 84.90\% & 53.94\% \\
MGDM          & 78.17\% & 78.70\% & 34.20\% & 34.80\% & 76.36\% & 8.70\%  & 30.13\% & 61.24\% & 84.90\% & 54.13\% \\
Blockwise     & \underline{79.08\%} & 76.57\% & 31.80\% & 34.20\% & 74.09\% & \underline{11.30\%} & 30.13\% & 61.53\% & \underline{85.75\%} & 53.83\% \\
DSFT          & 78.92\% & 79.15\% & \underline{35.00\%} & 36.00\% & 75.00\% & 6.58\%  & 29.24\% & \underline{61.56\%} & 84.64\% & 54.01\% \\
GIFT          & 77.71\% & 78.17\% & 31.00\% & 34.60\% & \underline{77.73\%} & 11.22\% & \underline{30.36}\% & 61.50\% & 85.41\% & 54.19\% \\
\midrule

\textbf{CSL}
& \textbf{79.83\%} & \textbf{80.52\%} & \textbf{36.40\%} & \textbf{38.60\%}
& \textbf{78.18\%} & \textbf{14.84\%} & \textbf{30.58\%} & \textbf{61.88\%} & \textbf{86.35\%}
& \textbf{56.35\%} \\

\rowcolor{blue!10} \textbf{$\Delta$ vs Base}
& +3.64\% & +0.16\% & +3.60\% & +3.20\% & +1.82\% & +3.66\% & +2.24\% & +0.67\% & +1.37\%
& +2.26\% \\

\rowcolor{red!10} \textbf{$\Delta$ vs SFT}
& +0.75\% & +1.44\% & +4.20\% & +2.00\% & +2.27\% & +3.56\% & +1.79\% & +0.48\% & +0.86\%
& +1.92\% \\

\midrule\midrule
\rowcolor{gray!20}
\multicolumn{11}{c}{\textbf{LLaDA-8B-Instruct}} \\
\midrule

\textbf{Base} & 78.17\% & \underline{80.97\%} & 32.80\% & 36.20\% & \underline{78.18\%} & 12.84\% & 29.24\% & 61.59\% & 85.15\% & 55.02\% \\
\textbf{SFT}            & 78.47\% & 79.23\% & \underline{33.80\%} & 36.80\% & 75.91\% & 11.38\% & 29.69\% & 61.78\% & 85.32\% & 54.71\% \\
\midrule

DiBT & 78.39\% & 78.01\% & 33.20\% & \underline{38.20\%} & 73.64\% & \underline{13.95\%} & 30.58\% & \underline{62.42\%} & 85.41\% & 54.88\% \\
MGDM          & 77.94\% & 79.23\% & 32.80\% & 35.20\% & 75.91\% & 10.12\% & 30.58\% & 61.97\% & 83.36\% & 54.12\% \\
Blockwise     & \textbf{80.14\%} & 79.38\% & 33.00\% & 36.60\% & \textbf{79.09\%} & 9.65\%  & 29.69\% & 62.26\% & \underline{85.49\%} & \underline{55.03\%} \\
DSFT          & 79.08\% & 78.09\% & \underline{33.80\%} & 35.60\% & 76.36\% & 13.28\% & 30.36\% & 61.66\% & 85.32\% & 54.84\% \\
GIFT          & 79.23\% & 78.47\% & 32.80\% & 34.60\% & 75.45\% & 12.78\% & \underline{31.03\%} & 62.26\% & \textbf{86.18\%} & 54.76\% \\
\midrule

\textbf{CSL}
& \underline{79.30\%} & \textbf{81.20\%} & \textbf{35.00\%} & \textbf{38.80\%}
& \textbf{79.09\%} & \textbf{14.01\%} & \textbf{31.25\%} & \textbf{62.45\%} & \underline{85.49\%}
& \textbf{56.29\%} \\

\rowcolor{blue!10} \textbf{$\Delta$ vs Base}
& +1.13\% & +0.23\% & +2.20\% & +2.60\% & +0.91\% & +1.17\% & +2.01\% & +0.80\% & +0.34\%
& +1.27\% \\

\rowcolor{red!10} \textbf{$\Delta$ vs SFT}
& +0.83\% & +1.97\% & +1.20\% & +2.00\% & +3.18\% & +2.63\% & +1.56\% & +0.61\% & +0.17\%
& +1.58\% \\

\bottomrule
\end{tabular}
\end{table*}

\begin{table}[ht]
\centering
\small
\setlength{\tabcolsep}{4pt}
\renewcommand{\arraystretch}{1.15}
\setlength{\tabcolsep}{3pt}
\caption{Performance on code generation tasks. Best and second-best results are \textbf{bold} and \underline{underlined}.}
\label{tab:code_results}

\begin{tabular}{lcccc|c}
\toprule

& \multicolumn{2}{c}{\textbf{HumanEval}} & \multicolumn{2}{c}{\textbf{MBPP}} & \\
\cmidrule(lr){2-3} \cmidrule(lr){4-5}

\textbf{Seq Len} & \textbf{256} & \textbf{512} & \textbf{256} & \textbf{512} & \textbf{Avg} \\
\midrule

\textbf{Instruct} & 29.88\% & 33.54\% & 38.13\% & 40.08\% & 35.41\% \\
\textbf{SFT}      & \underline{31.71\%} & 34.15\% & 39.69\% & 40.47\% & 36.51\% \\
\midrule

DiBT       & 25.61\% & 34.76\% & \underline{42.80\%} & 39.30\% & 35.62\% \\
MGDM       & 10.98\% & 10.37\% & 22.18\% & 16.34\% & 14.97\% \\
Blockwise  & 28.66\% & \underline{34.76\%} & 42.41\% & \underline{42.02\%} & \underline{36.96\%} \\
DSFT       & 25.61\% & 32.32\% & 36.19\% & 38.52\% & 33.16\% \\
GIFT       & 26.83\% & 32.93\% & 39.69\% & 39.30\% & 34.69\% \\
\midrule

\textbf{CSL}
& \textbf{32.32\%}
& \textbf{35.37\%}
& \textbf{43.19\%}
& \textbf{44.36\%}
& \textbf{38.81\%} \\

\rowcolor{blue!10} \textbf{$\Delta$ vs Base}
& +2.44\%
& +1.83\%
& +5.06\%
& +4.28\%
& +3.40\% \\

\rowcolor{red!10} \textbf{$\Delta$ vs SFT}
& +0.61\%
& +1.22\%
& +3.50\%
& +3.89\%
& +2.30\% \\

\bottomrule
\end{tabular}
\end{table}

\begin{table}[ht]
\centering
\footnotesize
\renewcommand{\arraystretch}{1.10}
\setlength{\tabcolsep}{2pt}

\caption{Ablation study on different extraction ratios $K$ and three extraction strategies.}
\label{tab:ablation_k}

\begin{tabular}{lcccc}
\toprule

& \multicolumn{2}{c}{\textbf{GSM8K}} & \multicolumn{2}{c}{\textbf{MATH}} \\
\cmidrule(lr){2-3}
\cmidrule(lr){4-5}

\textbf{Seq Len} & \textbf{256} & \textbf{512} & \textbf{256} & \textbf{512} \\
\midrule

\textbf{SFT (K=0)} 
& 79.08\% & 79.23\% & 32.20\% & 36.60\% \\
\midrule

Random (K=0.1L)
& 77.71\% & 78.54\% & 32.60\% & 35.20\% \\

One-step (K=0.1L)
& 77.94\% & 78.70\% & 33.20\% & 35.00\% \\

Step-by-step (K=0.1L)
& 78.47\% & 80.14\% & 34.00\% & 37.20\% \\
\midrule

Random (K=0.2L)
& 73.83\% & 76.12\% & 32.20\% & 34.20\% \\

One-step (K=0.2L)
& 75.21\% & 77.33\% & 32.80\% & 33.60\% \\

\rowcolor{red!10}
\textbf{Step-by-step (K=0.2L)}
& \textbf{79.83\%}
& \textbf{80.52\%}
& \textbf{36.40\%}
& \textbf{38.60\%} \\
\midrule

Random (K=0.3L)
& 72.25\% & 75.82\% & 28.20\% & 31.40\% \\

One-step (K=0.3L)
& 74.68\% & 75.13\% & 32.00\% & 33.60\% \\

Step-by-step (K=0.3L)
& 76.80\% & 79.08\% & 34.40\% & 35.20\% \\
\bottomrule

\end{tabular}
\end{table}

\begin{figure*}[t]
    \centering
    \begin{subfigure}[t]{0.49\textwidth}
        \centering
        \includegraphics[width=\linewidth]{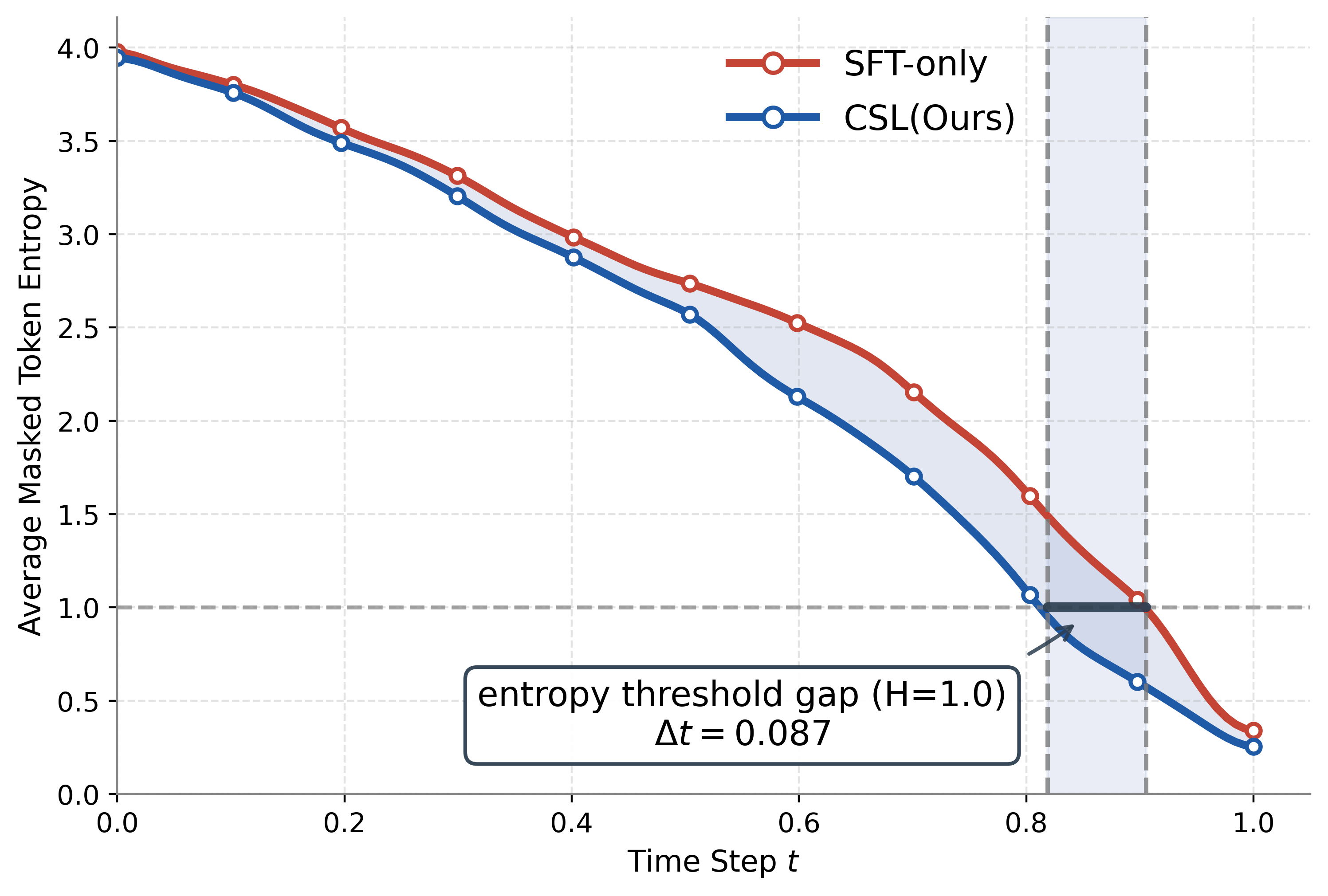}
        \caption{
        }
        \label{fig:gsm8k_right}
    \end{subfigure}
    \hfill
    \begin{subfigure}[t]{0.49\textwidth}
        \centering
        \includegraphics[width=\linewidth]{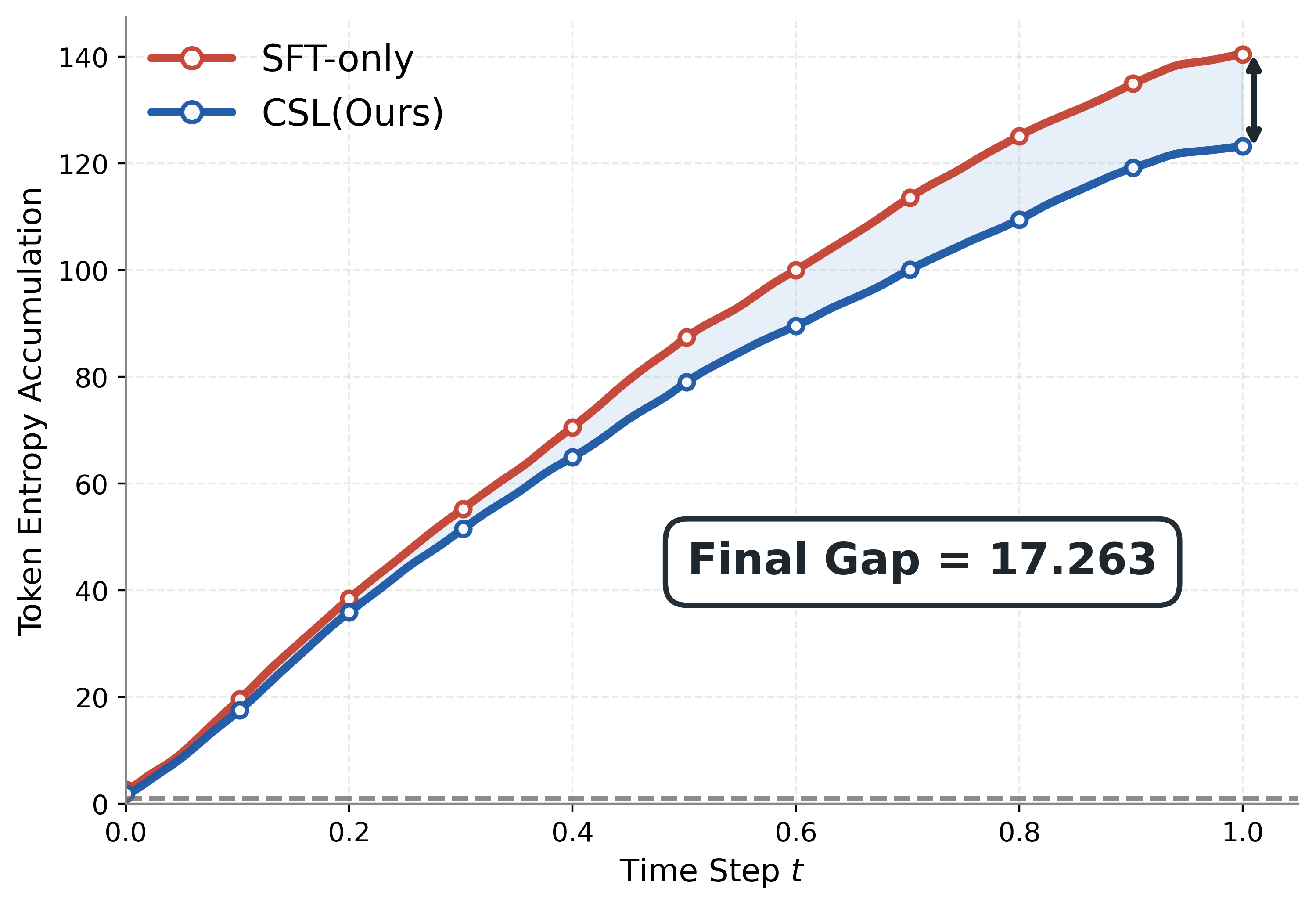}
        \caption{
        }
        \label{fig:accum}
    \end{subfigure}

    \caption{
    Comparison of entropy dynamics during decoding on GSM8K and MATH. (a) Evolution of the average masked-token entropy over decoding timesteps on GSM8K. When $t > 0.5$, CSL exhibits faster entropy decay. (b) Evolution of entropy accumulation over decoding timesteps on MATH. The gap increases with timesteps, reaching $17.263$ at the end.
    }
    \label{fig:entropy_dynamics}

\end{figure*}

\subsection{Main Results and Analysis}
\textbf{Math Reasoning Tasks.} We report the performance of different methods on seven mathematical reasoning benchmarks under two base models. For GSM8K and MATH-500, we further evaluate the models under different lengths. As shown in Table \ref{tab:results}, we observe that (1) CSL achieves the best overall performance across all benchmarks and outperforms all baselines under both base models, with average improvements of $1.92\%$ on LLaDA-8B-Instruct and $1.58\%$ on LLaDA-1.5; (2) CSL shows significant improvements on generation tasks such as GSM8K, MATH-500, and SAT, demonstrating stronger reasoning performance. In particular, it achieves a $4.20\%$ improvement on MATH-500 with a generation length of $256$; (3) Under the GSM8K-512 setting, most baseline methods suffer performance degradation after SFT, while CSL still maintains the best performance. These observations indicate that causal shortcuts provide efficient and accurate generation trajectories, and thus enhancing the reasoning capability of the model.

\begin{figure}[t]
\centering
\includegraphics[width=1.0\linewidth]{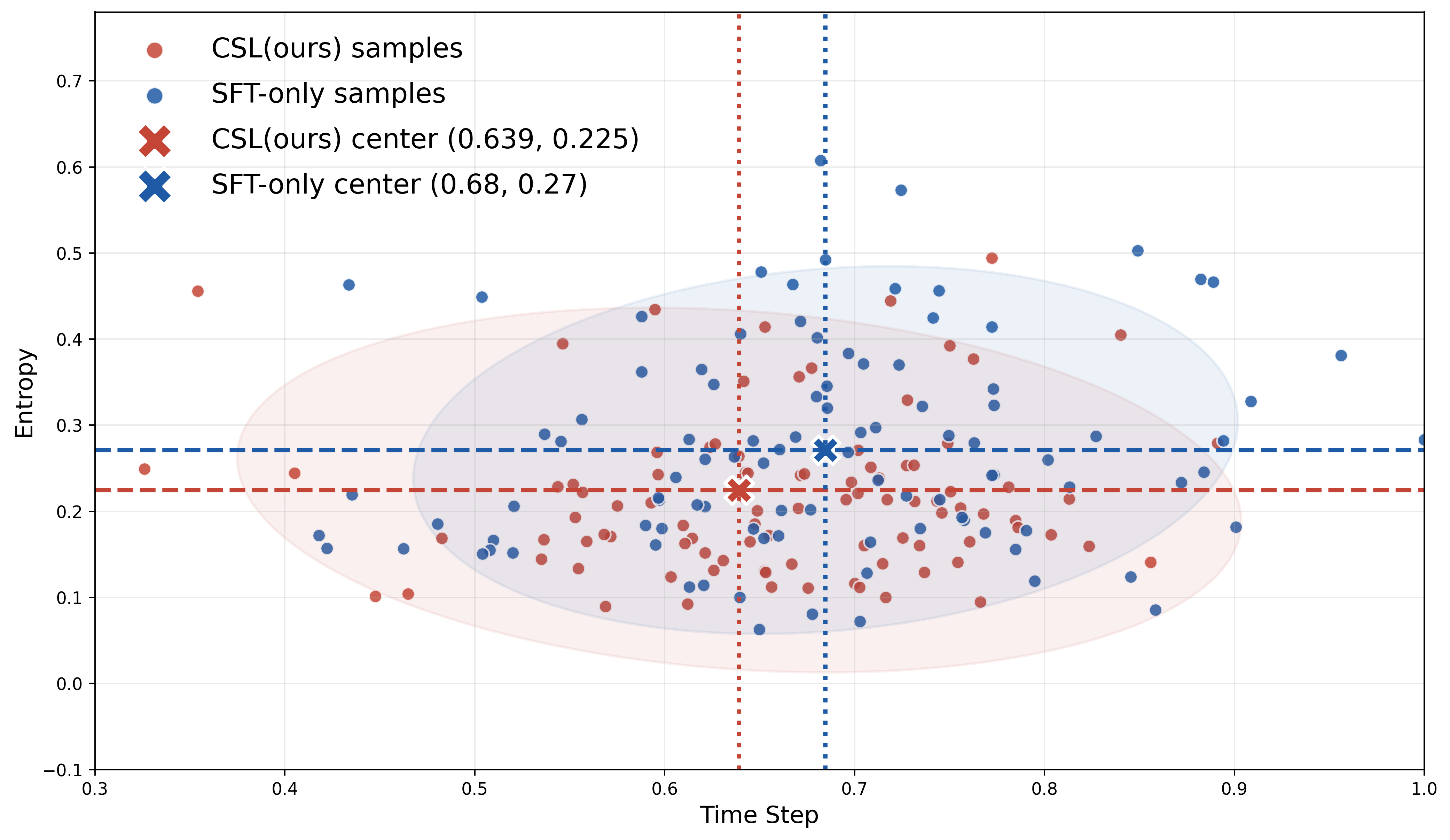}
\caption{
Scatter plot of token entropy and generation order for numbers and operators in the sequence. Y-axis denotes the average entropy and X-axis denotes average generation order.}
\label{fig:sandian}
\end{figure}

\noindent \textbf{Code Generation Tasks.} We report the performance on HumanEval and MBPP using LLaDA-8B-Instruct as the base model. As shown in Table \ref{tab:code_results}, (1) CSL outperforms all baselines and improves over SFT-only methods by an average of $2.30\%$, with gains of $3.50\%$ and $3.89\%$ on MBPP at lengths of $256$ and $512$, respectively; (2) The reweighting-based method MGDM suffers from severe performance collapse, highlighting the instability of reweighting-based training; (3) CSL shows larger gains under longer decoding lengths. While most baselines suffer from performance degradation due to error accumulation, CSL remains more stable and achieves higher accuracy. We further analyze the degradation issue in the ablation studies.

\renewcommand{\arraystretch}{1.00}
\setlength{\tabcolsep}{15pt}
\begin{table}[t]
\centering
\small
\setlength{\tabcolsep}{6pt}
\caption{Comparison of entropy decay between CSL and SFT on math tasks, measured by the average timestep $t$ at which entropy reaches the threshold of $1.0$. A smaller value indicates better performance.}
\label{tab:entropy_decay}

\begin{tabular}{l|cc|cc}
\toprule
\textbf{Method} 
& \multicolumn{2}{c|}{\textbf{GSM8K}} 
& \multicolumn{2}{c}{\textbf{MATH}} \\
\cmidrule(lr){2-3} \cmidrule(lr){4-5}
& Correct & All & Correct & All \\
\midrule
SFT & 0.901 & 0.916 & 0.922 & 0.930 \\
CSL & 0.814 & 0.869 & 0.867 & 0.891 \\
\textbf{$\Delta$} & \textbf{8.7\%} & \textbf{4.7\%} & \textbf{5.5\%} & \textbf{3.9\%} \\
\bottomrule
\end{tabular}
\end{table}

\renewcommand{\arraystretch}{1.00}
\setlength{\tabcolsep}{15pt}
\begin{table}[t!]
\centering
\small
\setlength{\tabcolsep}{6pt}
\caption{Comparison of entropy accumulation between CSL and SFT on math tasks. A smaller value indicates better performance.}
\label{tab:entropy_accum}

\begin{tabular}{l|cc|cc}
\toprule
\textbf{Method} 
& \multicolumn{2}{c|}{\textbf{GSM8K}} 
& \multicolumn{2}{c}{\textbf{MATH}} \\
\cmidrule(lr){2-3} \cmidrule(lr){4-5}
& 256 & 512 & 256 & 512 \\
\midrule
SFT & 57.78 & 94.62 & 84.90 & 138.4 \\
CSL & 51.36 & 85.32 & 78.72 & 121.1 \\
\textbf{$\Delta$} & \textbf{6.423} & \textbf{9.295} & \textbf{6.184} & \textbf{17.26} \\
\bottomrule
\end{tabular}
\end{table}

\subsection{Ablation Studies}\label{exction:ablation}

\textbf{Token Extraction Methods.}
To validate the effectiveness of token extraction, we conduct an ablation study under different set sizes $K \in \{0.0, 0.1L, 0.2L, 0.3L\}$, comparing three extraction methods on math tasks. As shown in Table \ref{tab:ablation_k}, both random and one-step extraction lead to performance degradation, suggesting that improper token extraction do disrupts inter-token dependencies and harms reasoning ability. The best performance is achieved at $K=0.2L$. When $K=0.1L$, causal shortcuts are insufficient to cover the entire trajectory, while $K=0.3L$ introduces excessive tokens that again degrade performance. We further discuss the selection of fixed and dynamic $K$ values under diverse training data in Appendix~\ref{appendix:kselection}.



\noindent \textbf{Entropy-Aware Analysis.}
We analyze CSL from three perspectives: (1) average entropy decay of masked tokens; We compare CSL with SFT on math tasks. As shown in Figure \ref{fig:gsm8k_right}, CSL exhibits faster entropy decay, especially for $t > 0.5$, and reaches the entropy threshold of $1.0$ earlier with a lead of $0.087$ timestep; (2) cumulative entropy of generated tokens. As shown in Figure \ref{fig:accum}, CSL maintains lower cumulative entropy throughout decoding, with an increasing gap over time. The performance discrepancies under various scenarios are detailed in Table~\ref{tab:entropy_decay} and \ref{tab:entropy_accum}; (3) Entropy and generation order of digits and operators: Digits and operators strongly reflect reasoning quality in mathematical reasoning tasks. As shown in Figure \ref{fig:sandian}, CSL shows lower entropy and earlier generation order for numbers and operators. 

In conclusion, CSL enables earlier and more confident decoding of causal shortcut tokens, leading to faster convergence and reduced error accumulation in long-sequence reasoning. Additional ablation studies are provided in Appendix \ref{appendix:ablation}.

\section{Conclusion}

To address the limited capability of DLMs in exploring effective generation trajectories, we propose Causal Shortcut Learning (CSL), a simple framework that leverages causal shortcuts to guide reasoning-oriented training. Extensive experiments on mathematical reasoning and code generation benchmarks demonstrate that CSL consistently improves accuracy over baselines. Further analysis shows that CSL improves the efficiency and stability of generation, while reducing error accumulation during long-sequence decoding. Overall, CSL provides an effective approach for enhancing reasoning in DLMs via causal shortcut learning.

\section*{Limitations}

Although our experiments are limited to 8B-scale models due to computational constraints, our method is fundamentally independent of model scale and architecture. We therefore expect it to scale effectively to larger models and further benefit from the stronger reasoning capabilities of more powerful DLMs.

Exploring proxy signals for token importance from internal representations, such as hidden states or attention patterns, is another promising direction. This would move beyond preprocessing-based importance scores and enable adaptive focus on informative tokens to enhance reasoning.

CSL represents an important exploration of importance-aware training in DLMs. We believe CSL highlights a fundamental challenge in DLM training: how to efficiently leverage data and training dynamics to better allocate learning capacity, making importance-aware training mechanisms an important direction for future research.

\section*{Ethical Considerations}

This work does not involve human subjects, personal data, or sensitive information, and all experiments are conducted on publicly available datasets. The proposed method is intended for research purposes only and may inherit biases from pretrained models. A potential risk is that it could be misused to inject illegal or unethical information into models; therefore, users should ensure compliance with relevant laws and ethical guidelines when applying this approach. It should not be used in safety-critical or high-stakes applications without proper validation and oversight.

\section*{Acknowledgements}
This work was supported in part by the National Key Research and Development Program of China (2024YFE0203700) and "Pioneer" and "Leading Goose" R\&D Program of Zhejiang (2025C02037). All opinions in this paper are those of the authors and donot necessarily reflect the views of the funding agencies.

\bibliography{custom}

\appendix

\section{Detailed Hyperparameter Settings}
\label{appendix:hyper}

In this section, we describe the detailed hyperparameter settings used in our experiments. All experiments are conducted on 4 NVIDIA A100 40G GPUs. 

\textbf{Training.} For models based on LLaDA-8B-Instruct, we train for $4$ epochs, while models based on LLaDA-1.5 are trained for $8$ epochs. For mathematical reasoning tasks, we use the Math-NoCoT-20k dataset for training, with the context length set to $2048$. For code generation tasks, we train on the OPC-SFT dataset with a context length of $1024$. We adopt LoRA for efficient fine-tuning, using a unified learning rate of $2\times10^{-4}$ and a LoRA rank of $8$. All training models in our experiments, including CSL and all baseline methods, follow the same training configuration described above. Additional hyperparameter settings specific to baseline methods are discussed in Appendix \ref{appendix:baselines}.

\textbf{Inference.} Different generation lengths are used according to task difficulty. For GSM8K, MATH-500, HumanEval, and MBPP, we evaluate with generation lengths of $\{256, 512\}$. For SAT, the generation length is set to $512$. For Sudoku, GPQA, MMLU-STEM, and ARC-C, the generation lengths are set to $128$, $64$, $32$, and $32$ respectively. The block generation size is uniformly fixed to $32$ for all experiments. All evaluation results are reported under the $0$-shot setting. We set the random seed to $42$ to ensure reproducibility.

More detailed hyperparameter settings can be found in our code repository.
















\begin{algorithm}[t]
\caption{Step-by-step Token Extraction}
\label{alg:token_extraction}
\begin{algorithmic}[1]
\REQUIRE Prompt $p$, initial sequence $x^{0}$,
target set size $K$, window size $D$
\ENSURE Selected index set $\mathcal{S}$

\STATE Construct initial sequence:
$x^{0} \leftarrow p + x^{0}$

\STATE Initialize $\mathcal{S} \leftarrow \emptyset$

\FOR{$k = 1$ to $K$}

    \STATE Obtain a unmasked candidate window
    $\mathcal{W}_k$ of length $D$
    
    \STATE Compute entropy:
    $\mathcal{H}(x^{k-1})$
    
    \FOR{each candidate position
    $i \in \mathcal{W}_k \setminus \mathcal{S}$}
    
        \STATE Unmask $x_i^{0}$ to obtain
        $x^{k-1}_{\setminus i}$
    
        \STATE Compute entropy:
        $\mathcal{H}(x^{k-1}_{\setminus i})$
    
        \STATE
        $\text{CMI}(x_i \mid x^{k-1})
        \leftarrow
        \mathcal{H}(x^{k-1})
        -
        \mathcal{H}(x^{k-1}_{\setminus i})$
    
    \ENDFOR
    
    \STATE
    $i^\star \leftarrow
    \arg\max_{i \in
    \mathcal{W}_k \setminus \mathcal{S}}
    \text{CMI}(x_i \mid x^{k-1})$
    
    \STATE
    $\mathcal{S}
    \leftarrow
    \mathcal{S} \cup \{i^\star\}$
    
    \STATE Unmask $x_{i^\star}$ in $x^{k-1}$
    to obtain $x^{k}$

\ENDFOR

\RETURN $\mathcal{S}$

\end{algorithmic}
\end{algorithm}

\section{Discussion on the Data Scaling of Step-by-Step Processing}
\label{appendix:scaling}

\noindent \textbf{Complexity of Extraction.} The step-by-step token extraction process requires $\mathcal{O}(K \cdot L)$ forward passes for entropy estimation, since each CMI computation requires one forward pass over the sequence, and all $L$ tokens are evaluated at each of the $K$ steps. For large-scale SFT datasets, step-by-step extract all samples is computationally expensive and time-consuming.

\noindent \textbf{CMI Score Model.} To address this issue, we introduce a CMI scoring model for efficient scaling. Specifically, a small-scale dataset is first used to construct supervised labels of CMI, where each sequence is annotated with token-level CMI scores. We then train a CMI Score Model built on a LLaDA backbone of the same scale. 

In the step-by-step extraction procedure, this method reduces the number of required forward passes per extraction from $O(K \cdot L)$ to $O(K)$, where $K$ denotes the number of causal tokens and $L$ is the sequence length.

A lightweight transformer block is inserted on top of the final hidden layer to predict a scalar score $s_i$ for each token, corresponding to its CMI value. The training objective is defined as:
$$
\mathcal{L}
=
\mathcal{L}_{\text{rank}}
+
\lambda \mathcal{L}_{\text{global}}
$$

Since we only focus on positions with maximum CMI values, we discard MSE loss and adopt ranking loss instead. The ranking loss enforces correct ordering of tokens according to CMI values by comparing pairwise relationships:
$$
\mathcal{L}_{\text{rank}}
=
\frac{1}{|\mathcal{P}|}
\sum_{(a,b)\in \mathcal{P}}
\mathrm{softplus}\left(
- \mathrm{sign}(\Delta_{ab}) \cdot (s_a - s_b)
\right)
$$
where $\Delta_{ab} = \mathrm{CMI}(x_a^t) - \mathrm{CMI}(x_b^t)$, and $s_i$ denotes the model’s predicted score $\hat{\text{CMI}}(x_i^t)$ at position $i$ in sequence $x$. The ranking objective encourages the model to assign higher scores to tokens with larger CMI values via relative comparisons. The global loss $\mathcal{L}_{\text{global}}$ is introduced to prevent excessive deviation in predicted scores across all tokens, defined as the squared difference between the mean predicted score and the mean target score:
$$
\mathcal{L}_{\text{global}} = \left( \frac{1}{L}\sum_{i=1}^{L} s_i - \frac{1}{L}\sum_{i=1}^{L} \hat{s}_i \right)^2.
$$

We adopt NDCG and Top-5\% Hit Rate as evaluation metrics to measure model performance. Normalized Discounted Cumulative Gain (NDCG) assesses the overall ranking quality, formulated as:
$$
\text{NDCG} = \frac{\text{DCG}}{\text{IDCG}},\quad \text{DCG}=\sum_{i=1}^n \frac{r_i}{\log_2(i+1)}
$$
where $r_i$ refers to the relevance score at position $i$, and $\text{IDCG}$ denotes the ideal discounted cumulative gain.

Top-5\% Hit Rate reflects the probability that the true maximum locates within the top 5\% ranked candidates, calculated by:
$$
\text{Top-5\% Hit Rate} = \frac{N_{\text{hit}}}{N_{\text{total}}}
$$
$N_{\text{hit}}$ is the quantity of samples with true maximum falling into the top 5\% range, and $N_{\text{total}}$ stands for the total number of test samples.

\begin{table}[ht]
\centering
\caption{Performance under Different $\lambda$ Values}
\label{tab:lambda_performance}
\begin{tabular}{lcccc}
\toprule
$\lambda$ & NDCG & Top 5\% Hit Rate \\
\midrule
0         &   82.5\% & 70.2\%                        \\
0.1       &   87.9\% & 76.7\%                \\
1.0       &   \textbf{92.1\%} & 85.4\%                \\
2.0       &   91.4\% & \textbf{86.1\%}                   \\
\bottomrule
\end{tabular}
\end{table}

\noindent \textbf{Ablation Studies} We first investigate the impact of $\lambda$ on evaluation metrics. As shown in \ref{tab:lambda_performance}, the scoring model achieves an NDCG score above $92.1\%$ and a Top-5\% Hit Rate of $85.4\%$ under $\lambda=1.0$, indicating that the learned ranking closely matches the ground-truth CMI ordering. This is attributed to the fact that the hidden states of the backbone model already encode rich representations, making CMI prediction effectively a projection from high-dimensional space to scalar information scores. This distillation strategy is applied consistently across both mathematical reasoning and code generation datasets.

\begin{table}[ht]
\centering
\small
\caption{Performance under Different Data Size}
\label{tab:data_size}
\begin{tabular}{lcccc}
\toprule
Data Size & NDCG & Top 5\% Hit Rate \\
\midrule
100       &   85.0\% & 74.3\% \\
500       &   90.3\% & 83.9\%                \\
1000      &   92.1\% & 85.4\%                \\ 
5000      &   92.6\% & 87.2\%                \\
\bottomrule
\end{tabular}
\end{table}

We further investigate the impact of data size on evaluation metrics. Based on the full dataset with around $20000$ samples, we test model performance on four subsets of different sizes: $\{100, 500, 1000, 2000\}$. The results are presented in Table~\ref{tab:data_size}. Insufficient data volume degrades ranking performance, while the marginal improvement gradually diminishes as data size grows. A subset of $1000$ samples achieves comparable performance to the large-scale dataset of $5000$ samples. 

\noindent \textbf{Why errors are negligible for exteacting causal shortcuts?} The major error stems from identifying the peak value within each cluster. When ranking quality is satisfactory, minor deviations only yield relatively high values around the true peak within the same cluster, exerting negligible influence on causal shortcut extraction.


\begin{figure*}[!t]
\centering
\begin{minipage}{\textwidth}
    \includegraphics[width=\textwidth]{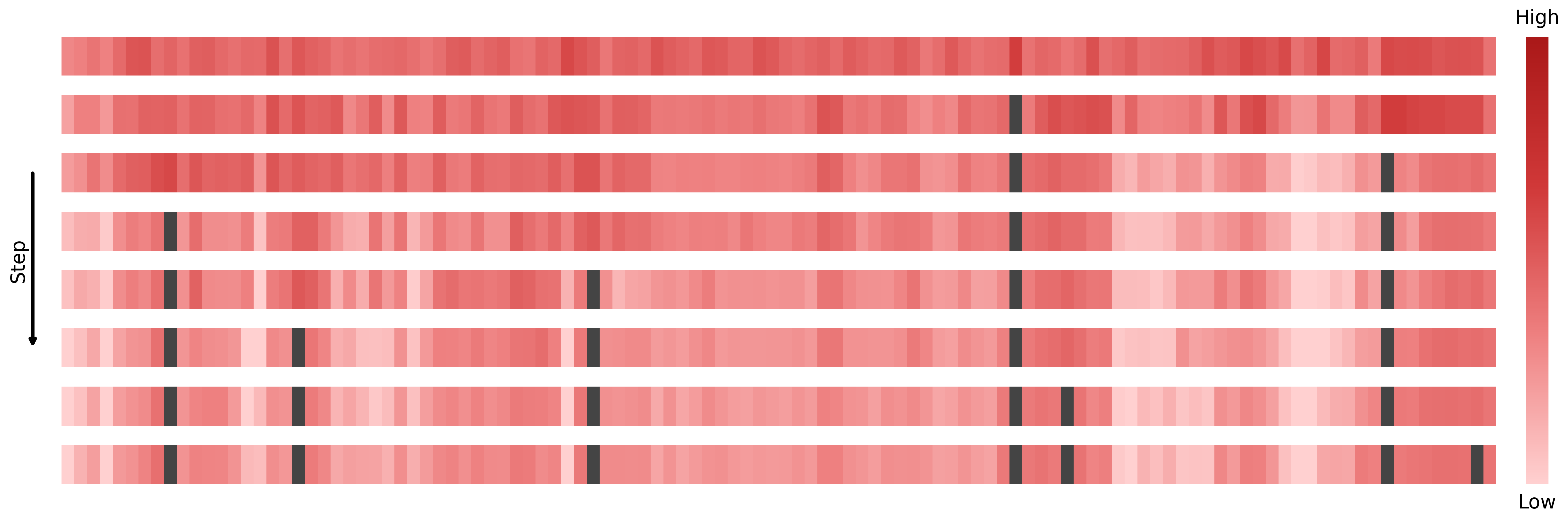}
\end{minipage}
\caption{Visual heatmap example of step-by-step extraction}
\label{fig:case}
\end{figure*}

\section{Theoretical Analysis of CMI and Step-by-step Extraction}\label{analysis}

We model the masked sequence at timestep $t$ as a union of $K$ latent dependency clusters:
$$
\mathcal{M}_t = \bigcup_{k=1}^{K} \mathcal{D}_k,
\quad
\mathcal{D}_i \cap \mathcal{D}_j = \emptyset.
$$

This decomposition is motivated by the observation that reasoning sequences exhibit a dominant local dependency structure: tokens within a reasoning unit (e.g., a step or sub-step) are strongly coupled, while interactions across different units are present but significantly weaker in magnitude. Therefore, the sequence can be effectively described by a cluster-wise organization of dependencies.

\paragraph{Intra-cluster dependence.}
Within each reasoning unit, tokens jointly encode the same semantic operation, leading to strong mutual dependencies:
$$
\begin{aligned}
I(x^i_t; x^j_t \mid x_t) \gg 0, 
\quad \forall x^i, x^j \in \mathcal{D}_k.
\end{aligned}
$$

While inter-cluster dependencies are not strictly zero in practice, they are significantly weaker than intra-cluster dependencies in reasoning sequences. Empirically, we observe that tokens within the same reasoning step exhibit substantially stronger mutual influence compared to tokens across different steps. Therefore, inter-cluster interactions can be treated as higher-order corrections, and the overall dependency structure is well-approximated by the dominance of intra-cluster relationships:
$$
\begin{aligned}
I(x^i_t; x^j_t \mid x_t) \gg I(x^i_t; x^r_t \mid x_t), \\
\quad
x^i_t,x^j_t \in \mathcal{D}_k,\; x^r \in \mathcal{D}_{k'},\; k \neq k'.
\end{aligned}
$$
This leads to a first-order approximation where CMI is primarily determined by intra-cluster contributions.

\paragraph{Entropy homogeneity.}
Due to shared contextual exposure under masking, tokens within the same cluster exhibit similar uncertainty levels:
$$
|H(x^i_t \mid x_t) - H(x^j_t \mid x_t)| \le \epsilon, \quad x^i_t, x^j_t \in \mathcal{D}_k.
$$

\paragraph{CMI decomposition.}
Under this dominant-structure approximation, conditional mutual information can be expressed as:
$$
\text{CMI}(x^i_t)
=
\sum_{k=1}^{K}
\sum_{x^j_t \in \mathcal{D}_k \setminus \{x^i_t\}}
I(x^i_t; x^j_t \mid x_t),
$$

and for $x_i \in \mathcal{D}_k$, we have the first-order approximation:
$$
\text{CMI}(x^i_t)
\approx
\sum_{x^j_t \in \mathcal{D}_k}
I(x^i_t; x^j_t \mid x_t),
$$
which shows that CMI is dominated by intra-cluster interactions, while inter-cluster effects are treated as higher-order corrections.

\paragraph{One-step extraction failure mode.}
The one-step selection rule
$$
\mathcal{S}_{\text{one}} = \operatorname{TopK}(\text{CMI})
$$
operates over a mixture of cluster-level CMI distributions. Since CMI is dominated by intra-cluster interactions, tokens within the same cluster exhibit comparable scores:
$$
\text{CMI}(x^i_t) \approx \text{CMI}(x^j_t), \quad x^i_t,x^j_t \in \mathcal{D}_k.
$$

This leads to redundant selection within high-density clusters:
$$
\mathbb{P}(\exists x^i_t,x^j_t \in \mathcal{D}_k \cap \mathcal{S}_{\text{one}}) \to 1,
$$
resulting in incomplete coverage of the global reasoning trajectory.

\begin{figure*}[t]
\centering
\small

\begin{subfigure}{0.48\textwidth}
    \centering
    \includegraphics[width=\linewidth]{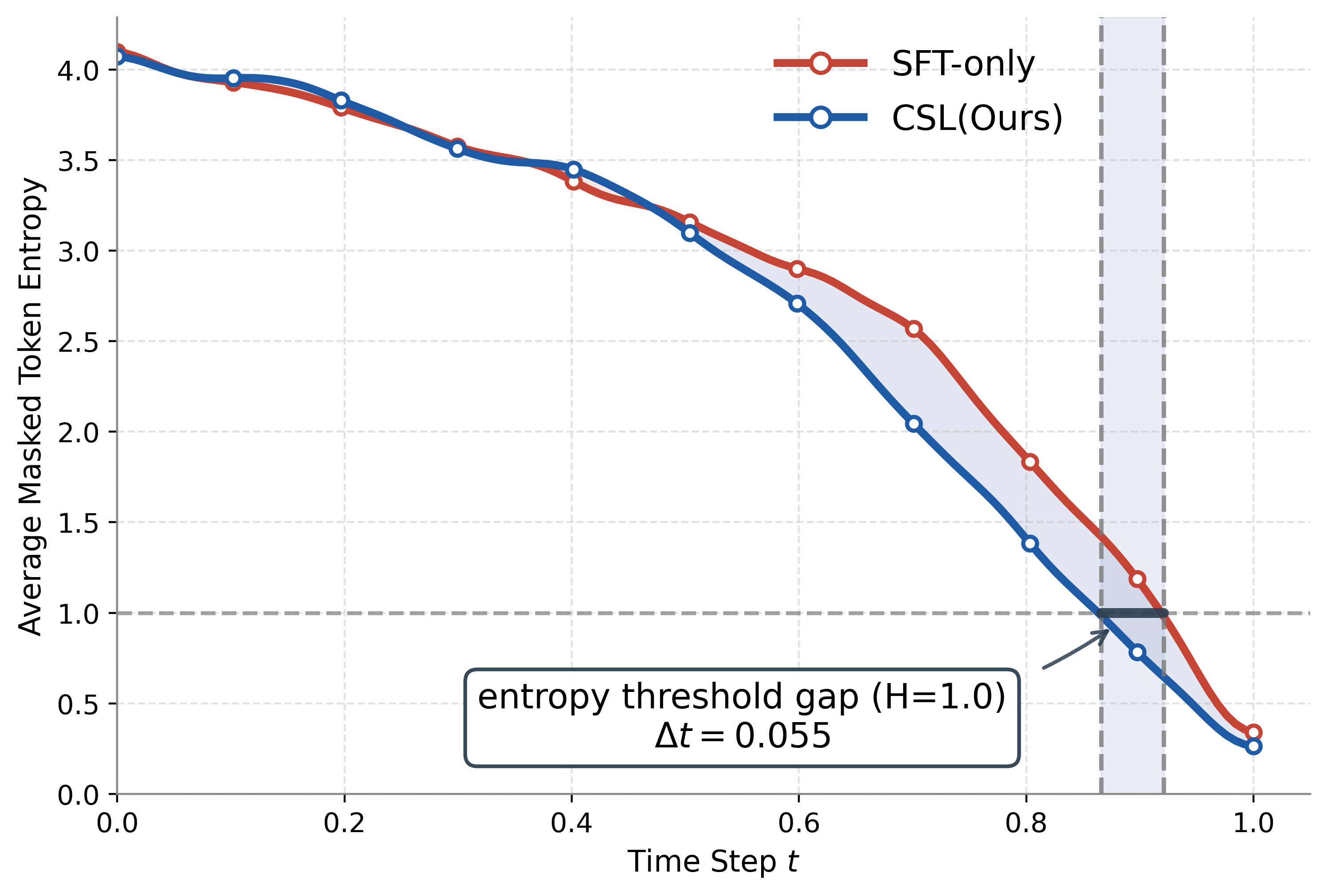}
    \caption{Math - Correct Sample}
\end{subfigure}
\hfill
\begin{subfigure}{0.48\textwidth}
    \centering
    \includegraphics[width=\linewidth]{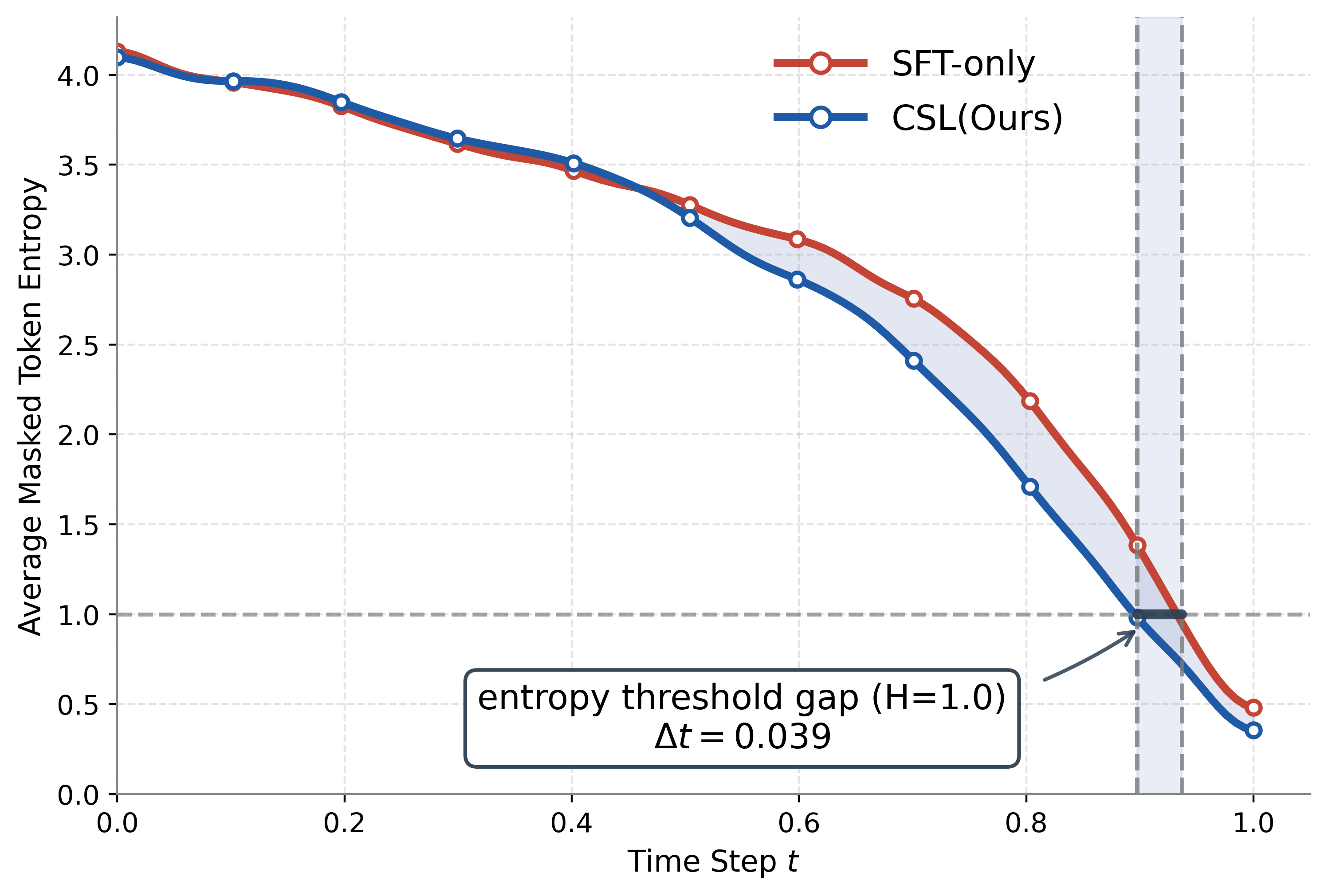}
    \caption{Math - All Sample}
\end{subfigure}

\vspace{0.3cm}

\begin{subfigure}{0.48\textwidth}
    \centering
    \includegraphics[width=\linewidth]{images/gsm8k_right.png}
    \caption{GSM8K - Correct Sample}
\end{subfigure}
\hfill
\begin{subfigure}{0.48\textwidth}
    \centering
    \includegraphics[width=\linewidth]{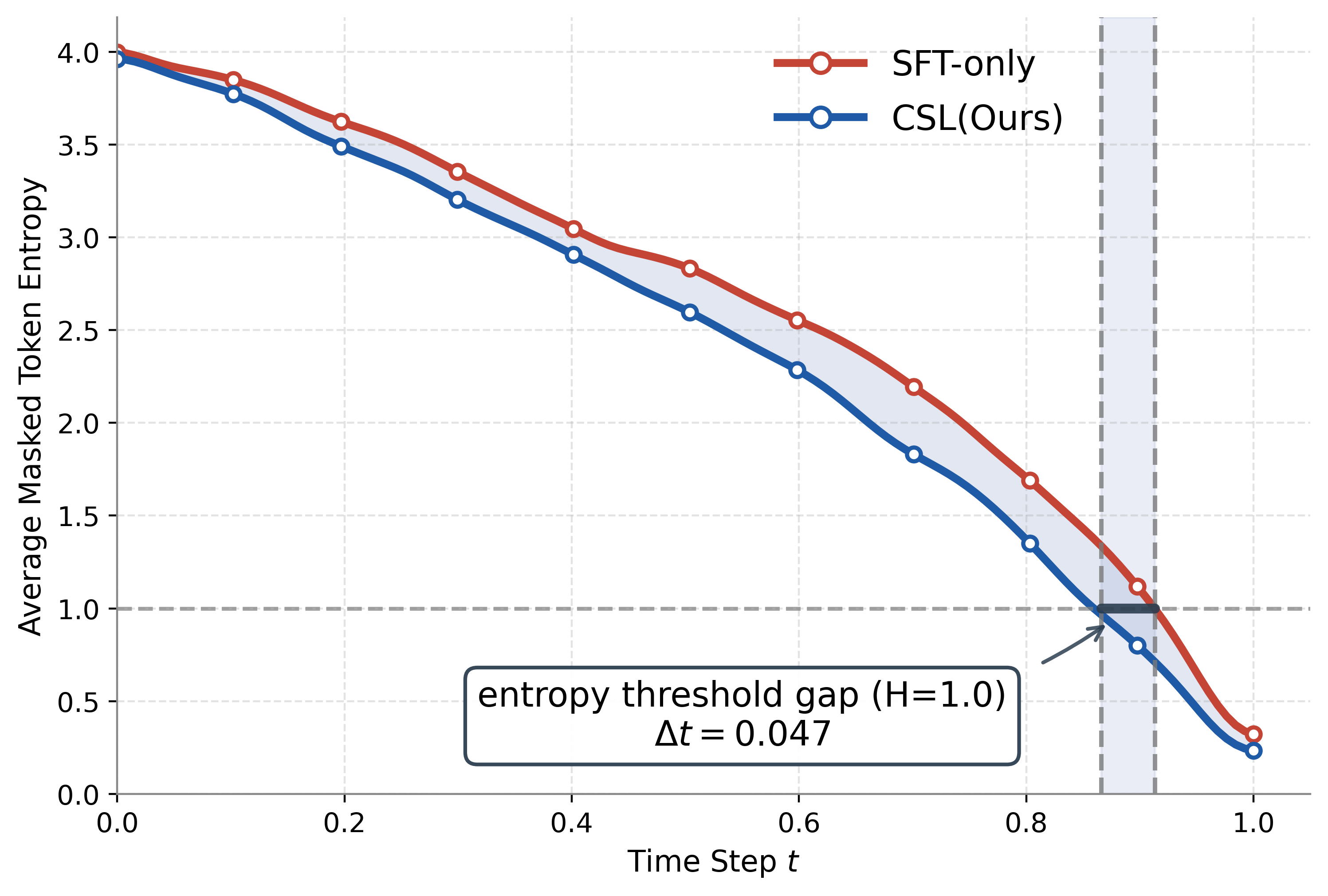}
    \caption{GSM8K - All Sample}
\end{subfigure}

\caption{
Comparison of entropy dynamics under four evaluation settings:  
(1) Math-500 (correct samples only),  
(2) Math-500 (all samples),  
(3) GSM8K (correct samples only),  
(4) GSM8K (all samples).  
}
\label{fig:entropy_4panel}

\end{figure*}

\begin{figure*}[t]
\centering
\small

\begin{subfigure}{0.48\textwidth}
    \centering
    \includegraphics[width=\linewidth]{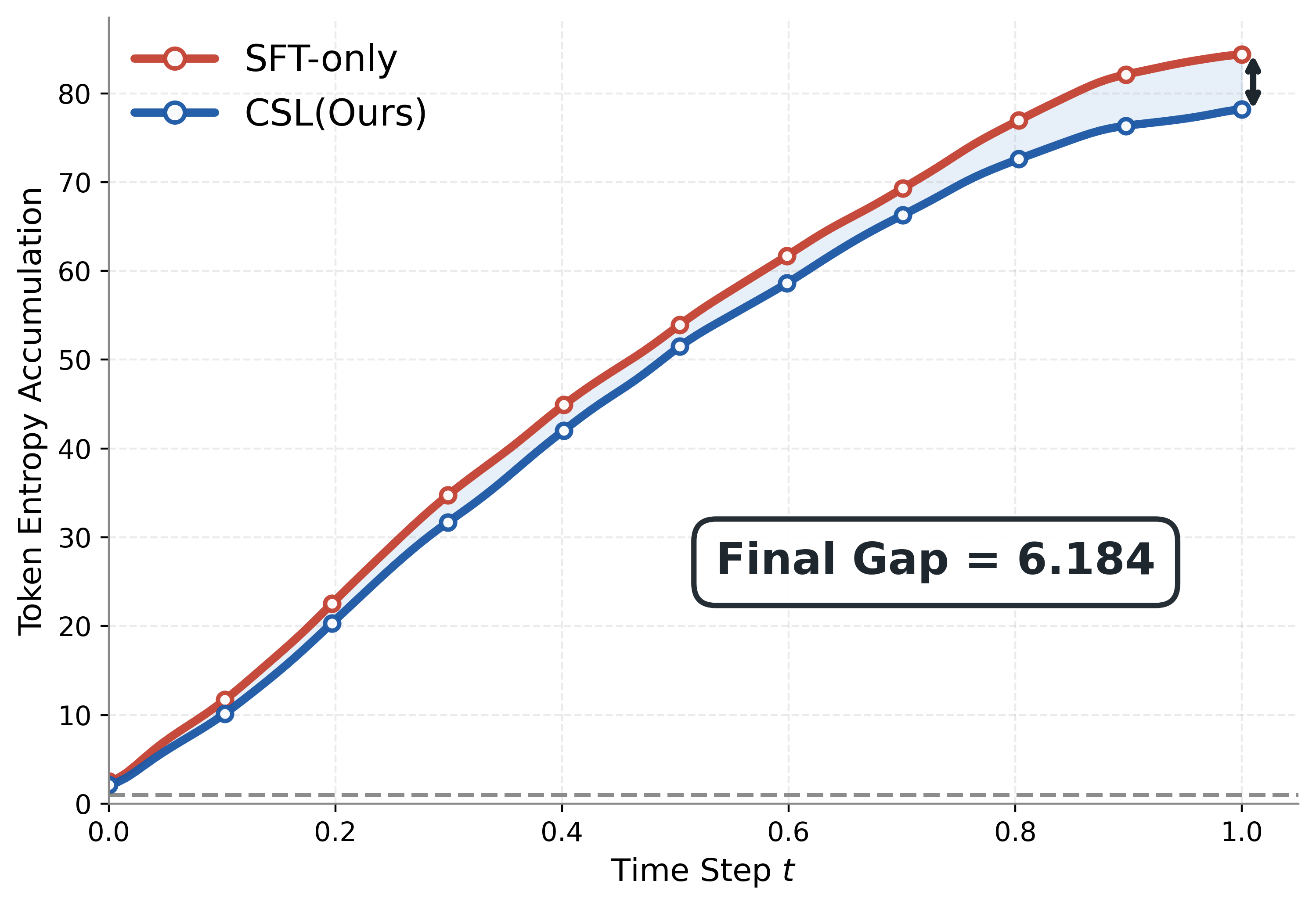}
    \caption{Math - Correct Sample}
\end{subfigure}
\hfill
\begin{subfigure}{0.48\textwidth}
    \centering
    \includegraphics[width=\linewidth]{images/math-512.png}
    \caption{Math - All Sample}
\end{subfigure}

\vspace{0.3cm}

\begin{subfigure}{0.48\textwidth}
    \centering
    \includegraphics[width=\linewidth]{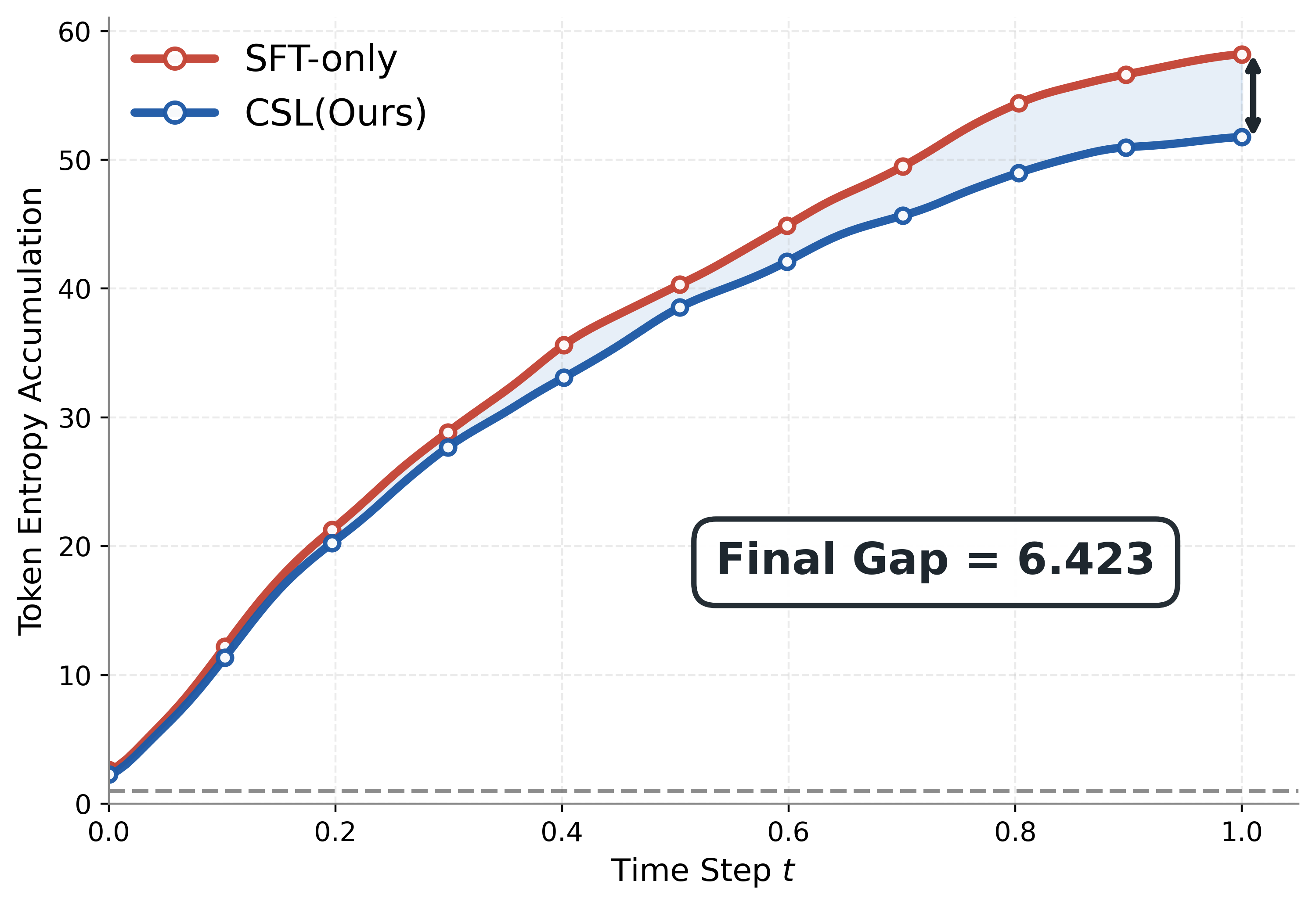}
    \caption{GSM8K - Correct Sample}
\end{subfigure}
\hfill
\begin{subfigure}{0.48\textwidth}
    \centering
    \includegraphics[width=\linewidth]{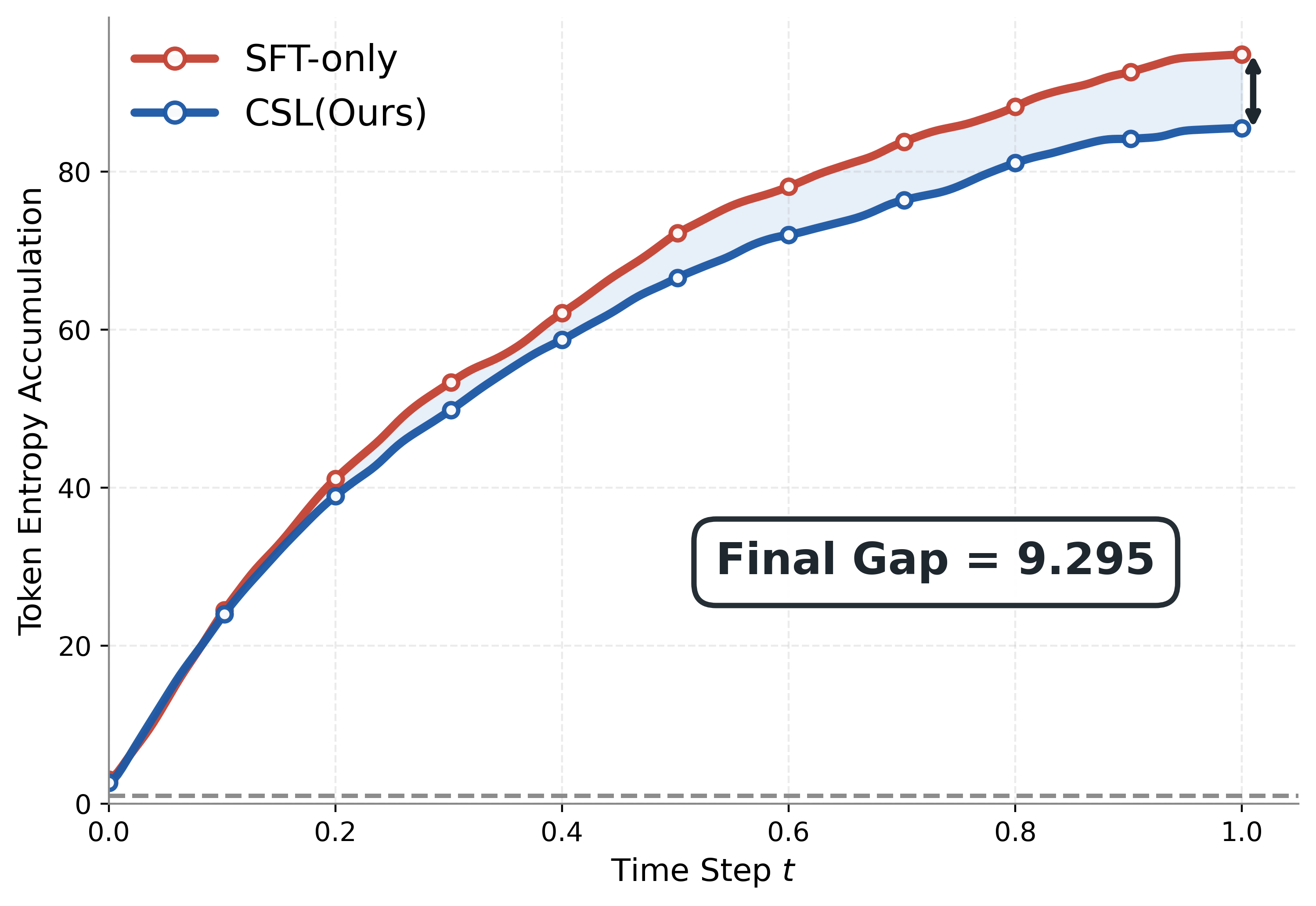}
    \caption{GSM8K - All Sample}
\end{subfigure}

\caption{
Comparison of entropy dynamics under four evaluation settings:  
(1) Math (generation length 256),  
(2) Math (generation length 512),  
(3) GSM8K (generation length 256),  
(4) GSM8K (generation length 512).  
}
\label{fig:accum_4panel}

\end{figure*}

\paragraph{Step-by-step extraction dynamics.}
Step-by-step extraction introduces an iterative suppression effect. After selecting a token $x_i \in \mathcal{D}_k$, information propagation reduces uncertainty within the same cluster:
$$
H(x^j_t \mid x^i_t, x_t) < H(x^j_t \mid x_t), \quad x^j_t \in \mathcal{D}_k.
$$

As a result, CMI within the same cluster is jointly suppressed:
$$
\text{CMI}^{(t+1)}(x^j_t) \le \text{CMI}^{(t)}(x^j_t), \quad x^j_t \in \mathcal{D}_k.
$$

This induces a cluster-level collapse effect:
$$
\max_{x^j_t \in \mathcal{D}_k} \text{CMI}^{(t+1)}(x^j_t)
\ll
\max_{x^r_t \notin \mathcal{D}_k} \text{CMI}^{(t+1)}(x^r_t).
$$

\paragraph{Cluster coverage property.}
As a result, step-by-step extraction behaves as a greedy cluster elimination process. After $K$ iterations:
$$
\mathbb{E}\big[|\mathcal{S} \cap \mathcal{D}_k|\big] \le 1,
\quad \forall k,
$$
which implies that the selected set $\mathcal{S}$ approximately covers distinct dependency clusters:
$$
|\mathcal{S}| = K \;\Rightarrow\; \mathcal{S} \approx \{\text{representatives of } \mathcal{D}_k\}.
$$

\paragraph{Interpretation.}
Therefore, CMI induces a latent clustering structure over tokens. As shown in Figure~\ref{fig:case}, step-by-step extraction performs a greedy cluster-wise uncertainty elimination process, which naturally leads to full trajectory coverage while avoiding redundant selection within the same reasoning region.

\section{Detailed Ablation Analysis}
\label{appendix:ablation}

This section provides additional entropy-aware ablation analyses that are not fully discussed in the main paper. We evaluate model behavior on the GSM8K and MATH-500 benchmarks under different settings, comparing CSL with the SFT-only baseline.

\textbf{Entropy Decay Curve.}  
The entropy decay curve reflects how quickly the model converges toward the final answer during generation. Evaluations are conducted on both the full test set and the intersection of correctly solved samples from the two models. As shown in Figure \ref{fig:entropy_4panel}, we observe that: (1) under all settings, the entropy curve of CSL consistently remains below that of SFT-only; and (2) the gap becomes significantly larger on correct samples. Using an average entropy threshold of $1.0$ to indicate convergence, CSL reaches the threshold $0.087$ timestep fraction earlier than SFT on correct samples. This behavior suggests that CSL enables the model to focus more effectively on causal shortcuts. As these tokens are generated with higher confidence and earlier decoding order, the model converges toward the correct answer more efficiently.

\textbf{Entropy Accumulation Curve.}
The entropy accumulation curve reflects the risk of error accumulation during generation, where larger accumulated entropy indicates higher uncertainty. Evaluations are conducted under different generation lengths. As shown in Figure \ref{fig:accum_4panel}, we observe that: (1) CSL consistently achieves lower accumulated entropy than SFT-only across all settings; and (2) the gap becomes larger under longer generation sequences, indicating that SFT-only suffers from more severe error accumulation in long-chain reasoning. This phenomenon is consistent with the core intuition of CSL: stronger confidence on causal shortcuts leads to more stable reasoning trajectories and lower uncertainty accumulation. In contrast, incorrect or unstable reasoning-guide tokens generated by SFT-only are more likely to mislead subsequent reasoning steps, resulting in degraded performance.


\textbf{Generation Order Analysis.}  
We analyze the entropy and decoding timestep of numerical and formula-related tokens on the MATH-500 benchmark. Lower entropy indicates higher confidence, while earlier decoding timesteps suggest stronger focus on reasoning-related content. For each sample, the average entropy and decoding timestep of reasoning-related tokens are computed and visualized as scatter plots. As shown in Figure \ref{fig:sandian}, CSL consistently achieves lower entropy and earlier decoding timesteps than SFT-only, indicating stronger confidence and earlier prioritization of reasoning-related tokens during generation.

\section{Analysis on CSL Training Stability}\label{appendix:training_loss}
Tokens with high CMI values usually possess high entropy, resulting in fluctuating loss values. We compare training loss behaviors between SFT and CSL, analyzing the average loss of all tokens and causal shortcut tokens respectively.

\begin{figure*}[t]
\centering
\small

\begin{subfigure}{0.48\textwidth}
    \centering
    \includegraphics[width=\linewidth, clip, trim=8 4 8 4]{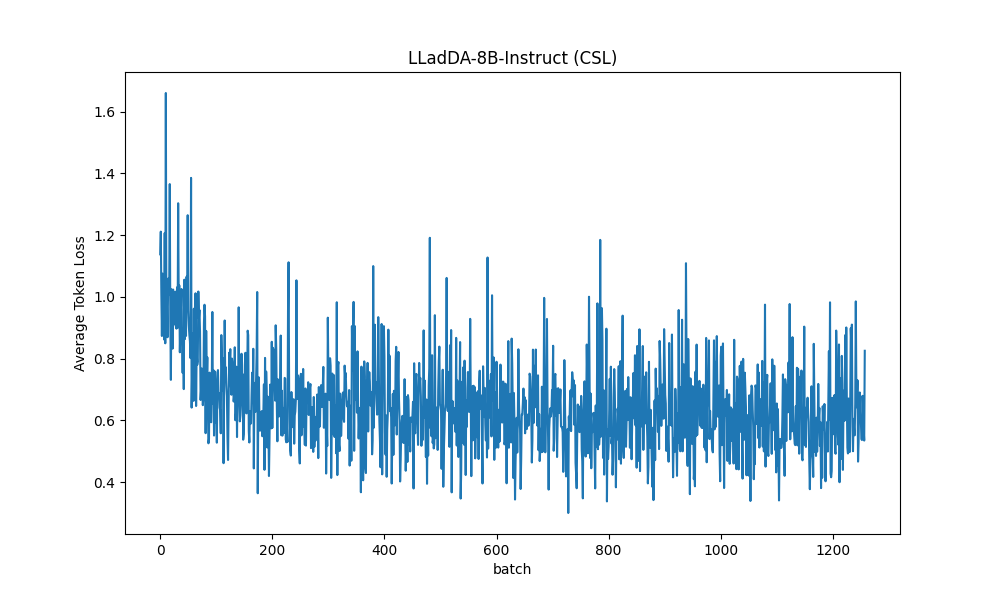}
    \caption{CSL training loss on LLaDA-8B-Instruct.}
\end{subfigure}
\hfill
\begin{subfigure}{0.48\textwidth}
    \centering
    \includegraphics[width=\linewidth, clip, trim=8 4 8 4]{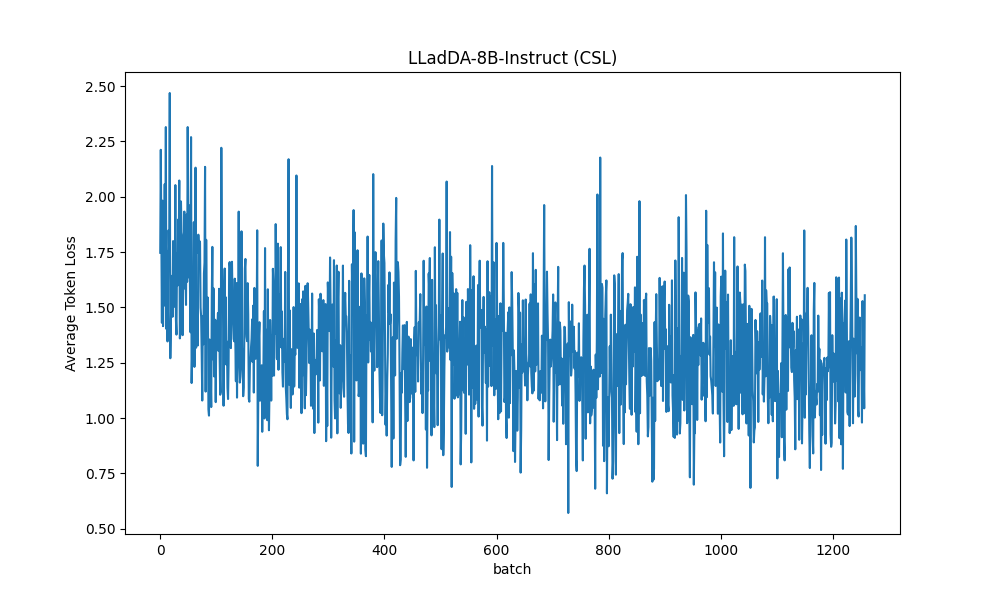}
    \caption{CSL causal shortcuts loss on LLaDA-8B-Instruct.}
\end{subfigure}

\vspace{0.3cm}

\begin{subfigure}{0.48\textwidth}
    \centering
    \includegraphics[width=\linewidth, clip, trim=8 4 8 4]{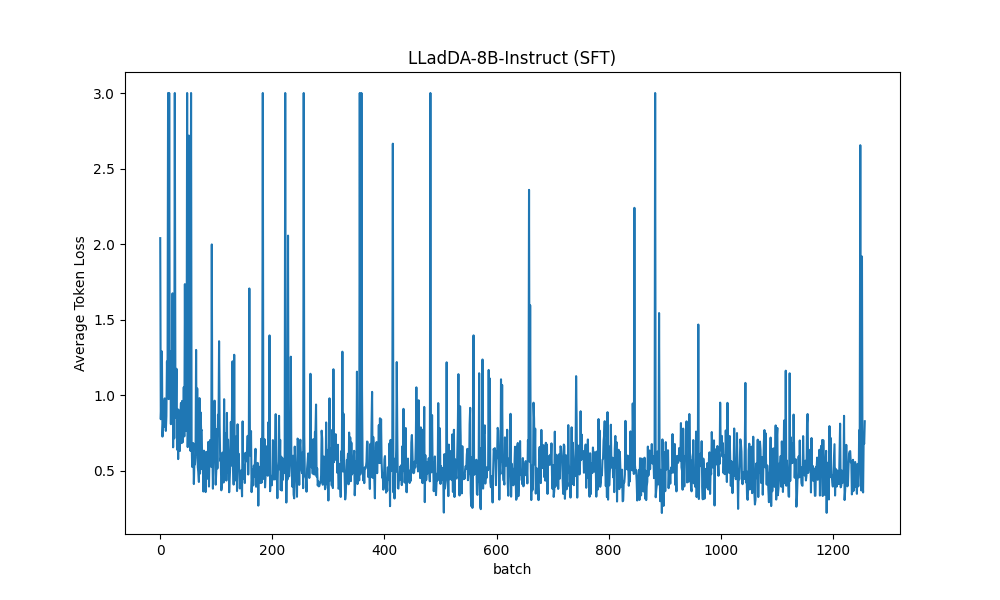}
    \caption{SFT training loss on LLaDA-8B-Instruct.}
\end{subfigure}
\hfill
\begin{subfigure}{0.48\textwidth}
    \centering
    \includegraphics[width=\linewidth, clip, trim=8 4 8 4]{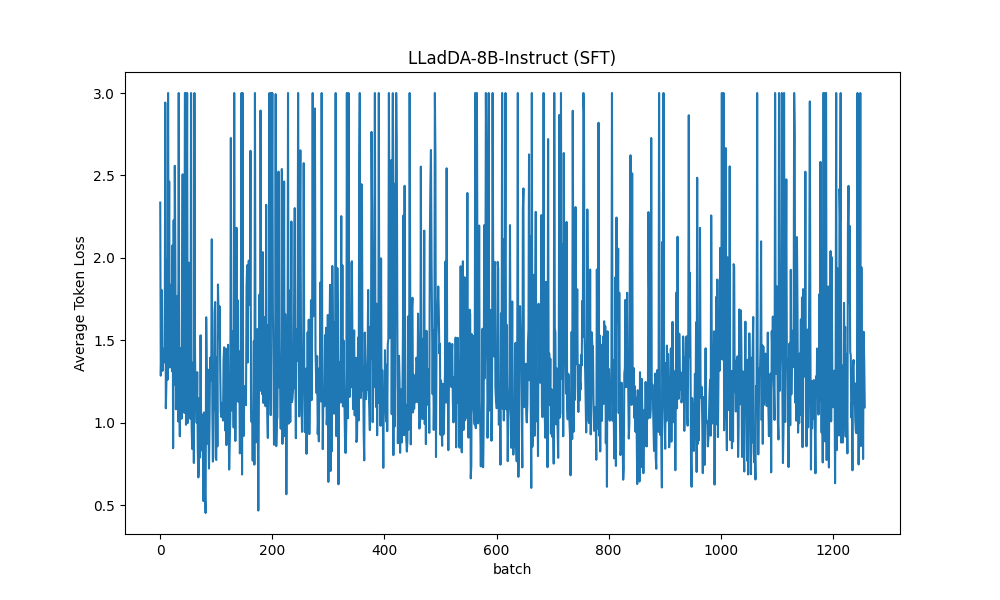}
    \caption{SFT causal shortcuts loss on LLaDA-8B-Instruct.}
\end{subfigure}

\begin{subfigure}{0.48\textwidth}
    \centering
    \includegraphics[width=\linewidth, clip, trim=8 4 8 4]{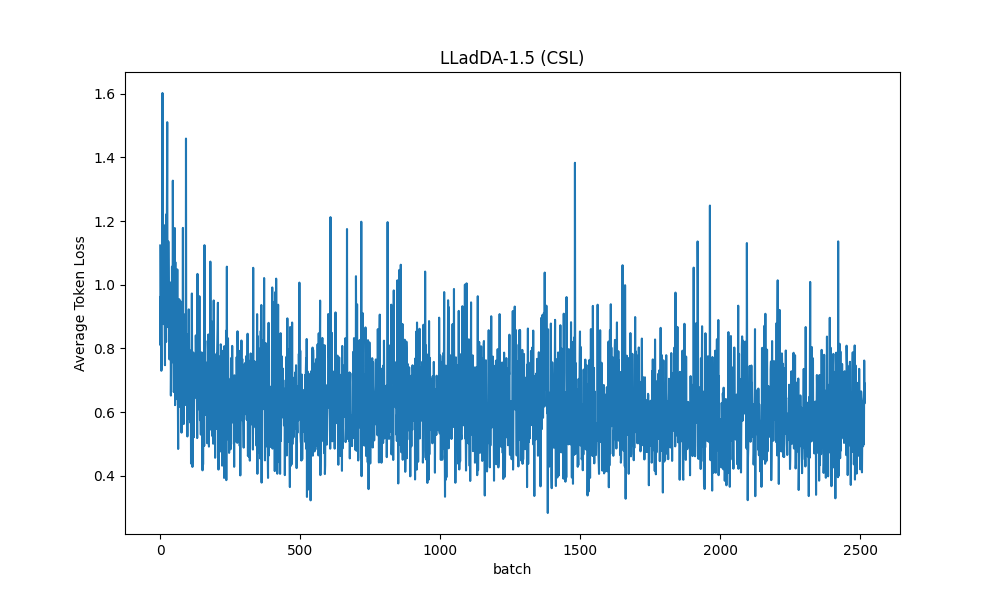}
    \caption{CSL training loss on LLaDA-1.5.}
\end{subfigure}
\hfill
\begin{subfigure}{0.48\textwidth}
    \centering
    \includegraphics[width=\linewidth, clip, trim=8 4 8 4]{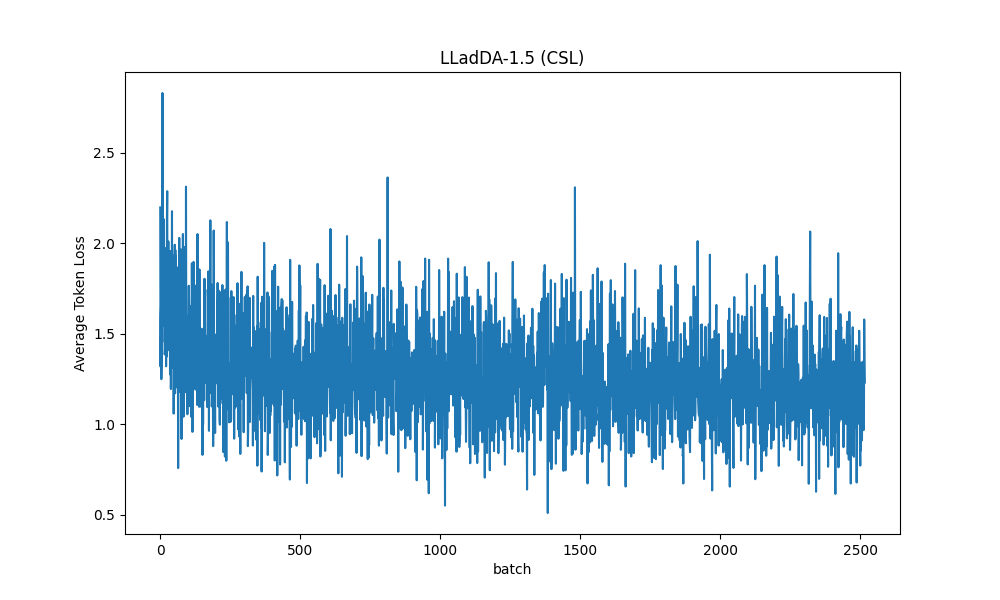}
    \caption{CSL causal shortcuts loss on LLaDA-1.5.}
\end{subfigure}

\vspace{0.3cm}

\begin{subfigure}{0.48\textwidth}
    \centering
    \includegraphics[width=\linewidth, clip, trim=8 4 8 4]{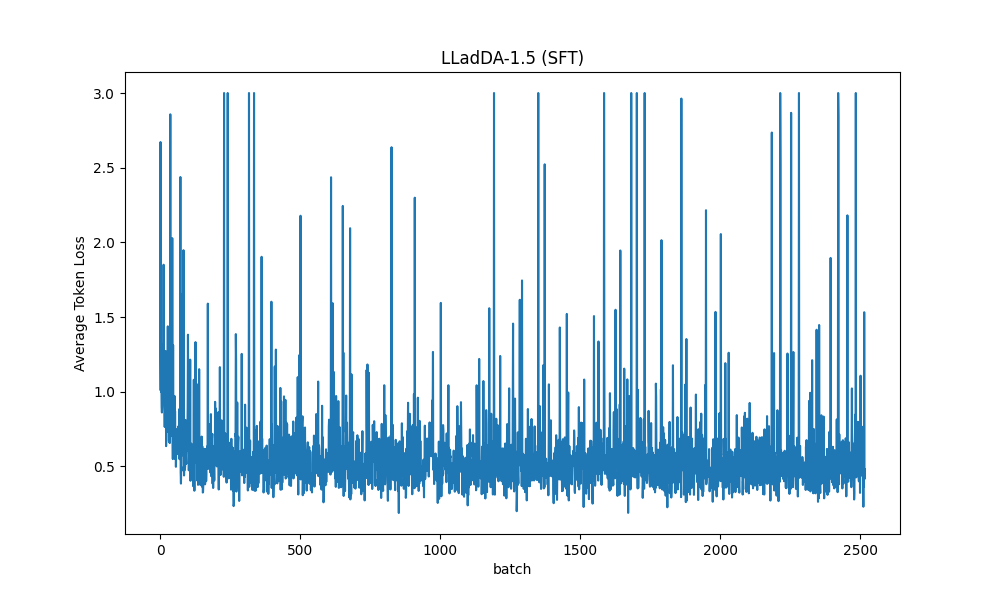}
    \caption{FT training loss LLaDA-1.5.}
\end{subfigure}
\hfill
\begin{subfigure}{0.48\textwidth}
    \centering
    \includegraphics[width=\linewidth, clip, trim=8 4 8 4]{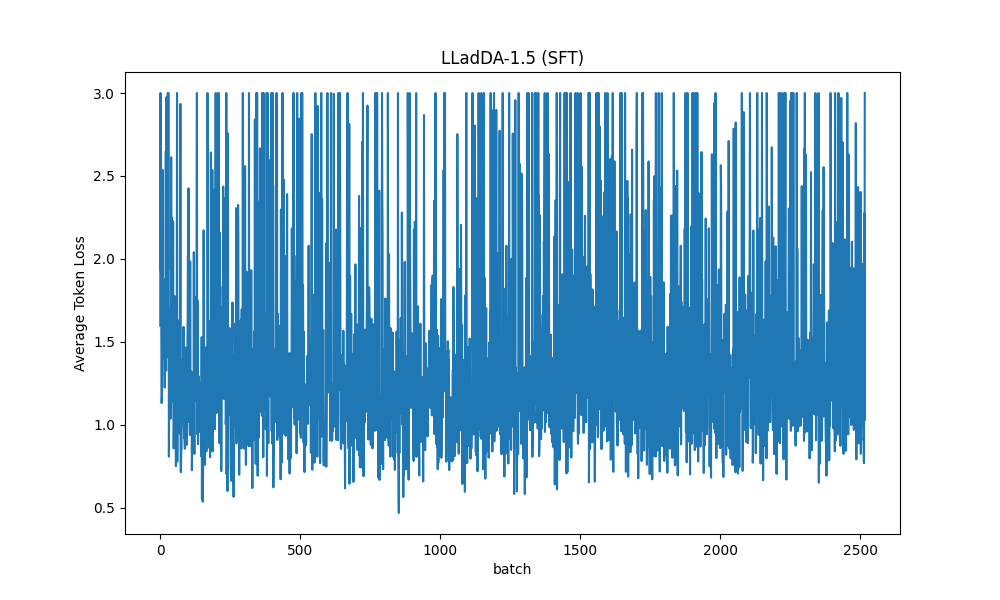}
    \caption{SFT causal shortcuts loss LLaDA-1.5.}
\end{subfigure}

\caption{Comparison of training loss and causal shortcuts loss between CSL and SFT on two base models.}
\label{fig:training_loss}

\end{figure*}

As illustrated in Figure~\ref{fig:training_loss}, the overall loss contains numerous outliers. Such extreme loss values emerge more frequently among causal shortcut tokens, which serve as the primary cause of training instability. In contrast, CSL training effectively reduces the occurrence of outliers. Frequent gradient updates optimized for causal shortcut tokens maintain stable overall training dynamics.

\begin{figure*}[t]   
\centering
\includegraphics[width=0.98\textwidth]{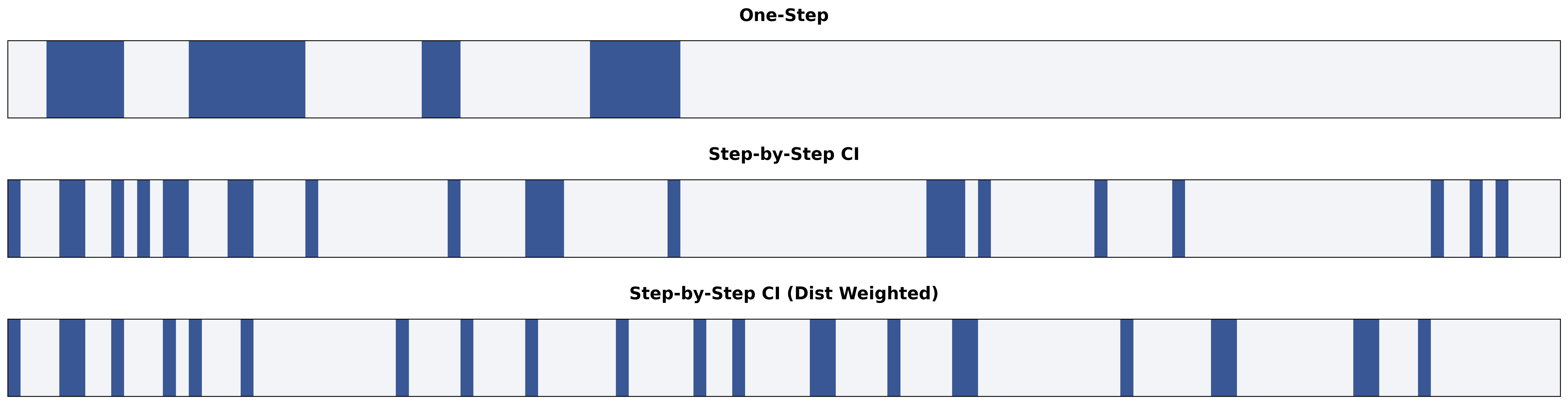}
\caption{Visual comparison of three different extraction strategies}
\label{fig:3extraction}
\end{figure*}

\section{Distance Weighted Conditional Mutual Information}

In the main paper, we use conditional mutual information (CMI) to quantify the information that a token contributes to the remaining unmasked sequence. The CMI score is defined as follows:
\begin{equation*}
\text{CMI}(x_t^i)
=
\frac{1}{|\mathcal{M}_t|-1}
\sum_{j \in \mathcal{M}_t \setminus \{i\}}
H(x_t^j)
-
H(x_t^j \mid x_0^i),
\end{equation*}
where $\mathcal{M}_t$ denotes the set of masked positions in the noisy sequence $x_t$.

Since we expect reasoning trajectories to cover the whole range, we attach greater importance to the local conditional mutual information ($\widetilde{\text{CMI}}$). We additionally introduce a distance-aware weighting mechanism. Specifically, we assign larger weights to nearby tokens when computing causal influence.
\begin{equation*}
\begin{aligned}
\widetilde{\text{CMI}}(x_t^i)
&=
\frac{1}{|\mathcal{M}_t|-1}
\sum_{j \in \mathcal{M}_t \setminus \{i\}}
w(i,j)
\\
&\quad \cdot
\left(
H(x_t^j)
-
H(x_t^j \mid x_i^0)
\right).
\end{aligned}
\end{equation*}
where $w(i,j)$ is a distance-based weighting function defined by a Gaussian kernel:
\begin{equation*}
w(i,j)
=
\exp\left(
-\frac{(i-j)^2}{2\sigma^2}
\right).
\end{equation*}

Here, $\sigma = L/4$, where $L$ denotes the sequence length. This weighting strategy encourages CMI to focus more on local reasoning structures and neighboring token dependencies.

\textbf{Comparison of Three Extraction Methods.}
We visualize the three extraction strategies. As shown in Figure~\ref{fig:3extraction}, obvious aggregation appears in the one-step method. The distance-weighted extraction focuses more on the local importance of individual tokens, enabling the token chain to distribute evenly across the entire sequence. This facilitates the guidance of valid training trajectories.

\begin{table}
\centering
\caption{Average entropy under different $K$ settings}
\label{tab:entropy_k}
\begin{tabular}{lccc}
\hline
$K$ & GSM8K & Math \\
\hline
$0.1L$ & 0.421 & 1.786 \\
$0.2L$ & 0.102 & 0.672 \\
$0.3L$ & 0.026 & 0.249 \\
\hline
\end{tabular}
\end{table}

\section{Selection of Causal Shortcut Set Size}\label{appendix:kselection}
According to the ablation study in Section \ref{exction:ablation}, the choice of $K$ substantially affects training performance. An overly small $K$ fails to achieve sufficient coverage of reasoning trajectories, while an excessively large $K$ causes continuous token accumulation and damages inherent dependencies.

\noindent \textbf{Data difficulty determines the set size.} In fact, the optimal size $K$ of the causal shortcut set depends on the difficulty of SFT data. As shown in Table~\ref{tab:entropy_k}, for GSM8K SFT samples, the sequence entropy drops to a sufficiently low level even when $K$ is set to $0.1L$ or smaller, leading to rapid model convergence. By contrast, sequences still maintain relatively high average entropy with $K=0.1L$ on MATH datasets. This indicates that simpler datasets require a smaller $K$, whereas more challenging tasks demand a larger value. In our experiments, we adopt the Math dataset with higher difficulty than GSM8K, where the setting $K=0.2L$ yields the optimal training performance. 

\noindent \textbf{Dynamic Set Size.} Furthermore, the utilized Math-CoT dataset presents balanced difficulty and stable CoT lengths, which justifies the adoption of a fixed $K$ during training. Nevertheless, for datasets with unbalanced difficulty and large variance in CoT length, such as mixed datasets, a dynamic $K$ strategy is necessary. A feasible solution is to set a predefined average entropy threshold for extraction such as ${1.0, 0.8}$, and adaptively determine $K$ according to the step where the average entropy drops below the threshold.

\section{Baseline Training Objectives}
\label{appendix:baselines}

In this section, we describe the implementation details of each baseline method to ensure fair comparison and reproducibility.

\textbf{DiffusionBERT (DiFT).\cite{db}} DiffusionBERT introduces an informativeness-aware noise schedule. The token-wise coefficient is defined as:
$$
\alpha_t^i
=
1
-
\frac{t}{T}
-
S(t)\cdot \tilde{H}(x_0^i),
$$
where
$$
S(t)
=
\lambda \sin\left(\frac{t\pi}{T}\right),
$$
and
$$
\tilde{H}(x_0^i)
=
1
-
\frac{\sum_{j=1}^{n} H(x_0^j)}
{n\,H(x_0^i)}.
$$

\noindent where $S(t)$ is a sinusoidal function controlling the influence of token informativeness across diffusion timesteps. The hyperparameter $\lambda$ is set to $1$ in all experiments. The resulting $\alpha_t^i$ is directly incorporated into the original NELBO objective to reweight the diffusion training process.

\textbf{MGDM.\cite{mgdm}} Multi-Granularity Diffusion Modeling (MGDM) introduces a hierarchical reweighting strategy that accounts for both sequence-level and token-level difficulty during diffusion training. The training objective is formulated as:
$$
\mathcal{L}_{\text{MGDM}}
=
\sum_{n=1}^{N}
\sum_{t=1}^{T}
w(t)\,v(x_{t,n})\,u(x_0, x_t, n; \theta),
$$
where the timestep weight is defined as $w(t) = \frac{\alpha'_t}{1 - \alpha_t}$. Under our setting $\alpha_t = 1 - t$, we obtain $w(t) = 1/t$. The token-level reweighting term is defined as:
$$
v(x_{t,n}) = \alpha \left(1 - \exp(-u(\cdot))\right)^{\beta}.
$$

\noindent where $\alpha$ controls the overall reweighting magnitude, while $\beta > 0$ emphasizes hard tokens and suppresses easy ones. In our experiments, we set $\alpha = 0.25$ and $\beta = 1$.

\textbf{Blockwise.\cite{blockwise}} Blockwise generation models semi-autoregressive decoding, where a response is generated as $M$ consecutive blocks $b^{(1)}, \dots, b^{(M)}$. The likelihood factorizes over blocks:
$$
p_\theta(x)
=
\prod_{a=1}^{M}
p_\theta\!\left(b^{(a)} \mid \text{context}(a), t=0\right),
$$
where $\text{context}(a)$ denotes the prefix up to block $a-1$. The training objective is to minimize the blockwise negative log-likelihood:
$$
\mathcal{L}_{\text{BW}}
=
\mathbb{E}_x
\left[
-\sum_{a=1}^{M}
\log p_\theta\!\left(b^{(a)} \mid \text{context}(a), t=0\right)
\right].
$$
In our experiments, we set the block size to $32$.

\textbf{DSFT.\cite{dsft}} DSFT adjusts the masking strategy and loss weighting from multiple perspectives to improve diffusion SFT.

Masking: Number-first masking additionally masks numerical tokens with probability $t / \text{num\_weight}$, where $\text{num\_weight}=0.5$. Span masking randomly applies contiguous spans with length sampled from $[2,5]$. Curriculum masking samples $t \sim \mathcal{U}(0,1)$ as the base masking rate.

Reweighting: Number-weighted loss assigns higher weights to numerical tokens during optimization. The training objective is:
$$
\mathcal{L}_{\text{DSFT}}
=
\frac{\sum_{i \in M} w_i \cdot \mathrm{CE}(f_\theta(x_t)_i, x_{0,i})}
{\sum_{i \in M} w_i},
$$

$$
w_i =
\begin{cases}
w_{\text{num}}, & \text{if } x_{0,i} \text{ is numerical} \\
1, & \text{otherwise}
\end{cases}
$$

\noindent where $w_{\text{num}} > 1$. In our experiments, we set $w_{\text{num}} = 2$.

\textbf{GIFT.\cite{gift}} GIFT introduces an importance-weighted diffusion SFT framework that adapts token-level masking based on token difficulty.

The token importance weight is computed from model prediction entropy:
$$
\beta_i = \sqrt{H(\mathrm{softmax}(z_i))},
$$
where $z_i = \mathrm{model}(\cdot \mid x, [M]).\mathrm{logits}$ denotes the logits at position $i$.

Given a token-specific masking rate $\beta_{x_i}$ and a reference rate $\beta_{\text{ref}}$, the masking probability at timestep $t$ is defined as:
$$
t_i = 1 - (1 - t)^{\frac{\beta_{x_i}}{\beta_{\text{ref}}}}.
$$

In our implementation, $\beta_{\text{ref}}$ is set to the mean of all $\beta$ values for numerical stability.

The final training objective is an importance-weighted supervised loss:
$$
\mathcal{L}_{\text{GIFT}}
=
- \sum_i
\mathbb{E}_{t_i}
\left[
\mathbb{I}(x_i^{t_i} = M)\cdot \frac{1}{t_i}
\log p_\theta(x_i^0 \mid x_t)
\right].
$$

\section{Use Of AI Assistants}
During the manuscript writing process, we used AI-assisted tools to help identify and correct syntactic and grammatical issues in the text. All suggested revisions were carefully reviewed and manually verified to ensure the accuracy, consistency, and integrity of the final content.

\end{document}